\pdfoutput=1

\documentclass[11pt]{article}

\usepackage[]{ACL2023}

\usepackage{times}
\usepackage{latexsym}

\usepackage[T1]{fontenc}

\usepackage[utf8]{inputenc}

\usepackage{microtype}

\usepackage{inconsolata}

\usepackage{algorithm}
\usepackage{algorithmic}
\usepackage{amssymb}
\usepackage{amsmath}
\usepackage{multirow}
\usepackage{graphicx}

\usepackage{longtable}
\usepackage[figuresright]{rotating}

\title{Exploring and Verbalizing Academic Ideas by Concept Co-occurrence}

\author{Yi Xu\textsuperscript{1},\ Shuqian Sheng\textsuperscript{1},\ Bo Xue\textsuperscript{1},\ Luoyi Fu\textsuperscript{1}\thanks{\quad  Luoyi Fu is the corresponding author.},\ Xinbing Wang\textsuperscript{1},\ Chenghu Zhou\textsuperscript{2} \\
        \textsuperscript{1}Shanghai Jiao Tong University, Shanghai, China \\
        \textsuperscript{2}IGSNRR, Chinese Academy of Sciences, Beijing, China\\
        \texttt{ \{yixu98, susisheng, sappho\_x, yiluofu, xwang8\}@sjtu.edu.cn }}

\begin{document}
\maketitle
\begin{abstract}
Researchers usually come up with new ideas only after thoroughly comprehending vast quantities of literature. The difficulty of this procedure is exacerbated by the fact that the number of academic publications is growing exponentially. In this study, we devise a framework based on concept co-occurrence for academic idea inspiration, which has been integrated into a research assistant system. From our perspective, the fusion of two concepts that co-occur in an academic paper can be regarded as an important way of the emergence of a new idea. We construct evolving concept graphs according to the co-occurrence relationship of concepts from 20 disciplines or topics. Then we design a temporal link prediction method based on masked language model to explore potential connections between different concepts. To verbalize the newly discovered connections, we also utilize the pretrained language model to generate a description of an idea based on a new data structure called co-occurrence citation quintuple. We evaluate our proposed system using both automatic metrics and human assessment. The results demonstrate that our system has broad prospects and can assist researchers in expediting the process of discovering new ideas.\footnote{The project is publicly available for research purpose \url{https://github.com/xyjigsaw/Kiscovery}.}
\end{abstract}

\section{Introduction}
Academic publications have witnessed the evolution and advancement of human civilization. In modern society, out-of-box and interdisciplinary scientific work can get more attention from science funders, industry, and the public~\cite{thurner2020role}, where a good idea is the cornerstone of academic research. However, for most researchers, it takes a lot of time to put forward new ideas. For one thing, the number of academic publications is increasing exponentially, and it is difficult for an independent researcher to understand these papers thoroughly. Besides, researchers often focus on their specialized but narrow fields, which makes it a challenge to discover underlying connections beyond their familiar areas~\cite{lahav2022search, Krenn2020PredictingRT}. In this work, our purpose is to unveil the profound connections between different academic concepts and ignite researchers' exploration of potential academic ideas while expediting the research process. The two primary goals are \textbf{idea exploration and verbalization}.


For the first goal, we need to understand how new ideas originate. Generally speaking, the emergence of a simple idea is often formed by the interaction between two different concepts rather than from scratch. For example, the combination of \textit{convolution} and \textit{graph neural network} contributes to \textit{graph convolutional network}~\cite{kipf2017semisupervised}. This understanding of idea as connection and combination inspires us to model the process of idea exploration as a link prediction task based on the \textit{evolving co-occurrence graph} of concepts. Such graphs are constructed according to the co-occurrence relationship of concepts in the papers published in different years. \textbf{It should be highlighted that there exist numerous factors leading to new ideas in the real world. We provide a possible way as a preliminary exploration.} 

The second goal, idea verbalization, is carried out after idea exploration to generate fluent and reasonable texts describing an idea, which usually comprises new contents derived from the combination of two different concepts. We retrieve sentences pertaining to concepts from existing publications and then verbalize ideas using the technique of natural language generation. Specifically, We propose a new data structure called \textit{co-occurrence citation quintuple} (Figure~\ref{fig:quintuple}), which stores two concepts, their corresponding sentences of papers, and idea texts. The definition is given in section~\ref{sec:data}. The quintuple is an extension of edges in the evolving concept co-occurrence graph and indicates where an idea comes from. We use such quintuples to train a sequence-to-sequence text generation model.

In our application scenario, there are various types of disciplines. Each of them has distinct characteristics and concepts. Existing methods of link prediction and text generation~\cite{yao2019kg,wang2019paperrobot,Krenn2020PredictingRT,pareja2020evolvegcn,da2022multi} are mostly trained on one dataset by optimizing a set of parameters. Owing to the fact that different datasets require specific training configurations and hyper-parameters, such models cannot be transferred to other datasets. Particularly, link prediction models need to set the scale of graphs before training, such as the number of nodes. Moreover, in the field of natural language generation, some works~\cite{wang2019paperrobot,yu2022survey} tend to construct domain knowledge bases as external information to generate texts. However, building large knowledge bases for each discipline takes tremendous resources, which is unrealistic. To this end, it is preferable to design general and informative models which can be applied to numerous disciplines.

Thanks to the abundant training corpus of pretrained language models (PLMs) such as BERT~\cite{devlin2018bert}, T5~\cite{raffel2020exploring}, BART~\cite{lewis2020bart}, and GPT~\cite{radford2018improving}, PLM can be regarded as an implicit knowledge graph~\cite{petroni2019language, wang2020language}, which has the ability of extrapolation. In this work, we integrate the whole academic information into the same representation space by leveraging the capability of PLM to break through disciplinary barriers. For idea exploration, we devise a PLM-based link prediction method, which only needs to train one set of model parameters. For idea verbalization, we use another sequence-to-sequence-based PLM endowed with academic knowledge from millions of highly-cited papers via unsupervised denoising training. Subsequently, we re-train the denoised PLM with co-occurrence citation quintuples in a supervised way. Our contributions are summarized as follows:
\begin{itemize}
    \item \textbf{New insights}: we transform the idea generation into two sequential sub-tasks: temporal link prediction and idea verbalization. The former aims to model and predict potential concept connections, while the latter involves expressing these new connections in natural language.
    \item \textbf{Publicly-released datasets}: we construct 240 evolving concept co-occurrence graphs with 20 high-level disciplines and topics. Each of them includes 23 annual snapshots ranging from 2000 to 2022. For idea verbalization, we propose a new data structure known as the co-occurrence citation quintuple that reveals how ideas appear. We curate nearly 10K high-quality co-occurrence citation quintuples, which originate from 29M papers with high citations.
    \item \textbf{General system for all disciplines}: we design a novel temporal link prediction method and train an idea verbalization model with a large number of academic papers. The two modules are integrated into a system to serve researchers from different fields. Note that the system updates the latest papers to encourage new ideas sustainably. Users are free to enter any academic query.
    \item \textbf{Systematic experiments}: we conduct extensive experiments, including automatic metrics and human assessment, to evaluate the performance of our link prediction method and idea verbalization model. The results show that our system has a promising prospect of helping researchers discover new ideas. 
\end{itemize}

\section{Preliminaries}
\subsection{Evolving Concept Co-occurrence Graph}
Given a concept set $C=\{c_i\}_{i=1}^{N}$ consisting of $N$ concepts and a paper corpus $P=\{p_j\}_{j=1}^{M}$ consisting of $M$ papers, let $C_p \subset C$ denote the set of concepts paper $p \in P$ contains. When concepts $c_u$ and $c_v$ ($c_u \neq c_v$) occur together in the same paper $p$ at the same time, i.e., $c_u \in C_p, c_v \in C_p$, it is considered that $c_u$ and $c_v$ co-occur, that is, there is a connection between the two concepts. Let $\mathcal{A} \in \mathbb{R}^{N \times N}$ represent the co-occurrence matrix of any two concepts, which is defined as follows:

\begin{equation}
    \mathcal{A}(c_u, c_v)=\left\{
            \begin{aligned}
                &1,\quad \exists p,\enspace c_u \in C_p, c_v \in C_p\\
                &0,\quad otherwise
            \end{aligned}
        \right.
\end{equation}

A concept co-occurrence graph is a pair $\mathcal{G}=(C, E)$, where $C$ is a set of concepts, and $E$ is a set of edges representing the co-occurrence relationship between concepts. The co-occurrence matrix $\mathcal{A}$ is the adjacent matrix of $\mathcal{G}$. Let $G=\{\mathcal{G}_{t}\}_{t=T_s}^{T_e}$ denote a set of concept co-occurrence graphs at different times ranging from $T_s$ to $T_e$, $\mathcal{A}_{t}$ represent the adjacent matrix of $\mathcal{G}_{t}$. We call $G$ \textit{evolving concept co-occurrence graph}. Similar to citation network, $G$ is a strictly evolving network~\cite{skarding2021foundations} where the connection of concepts has infinite duration. This implies that the edges in $G$ \textbf{never disappear}. Exploring ideas aims to predict future co-occurrence relations in $G$.

\begin{figure}[t]
    \centering
    \includegraphics[width=1.0\linewidth]{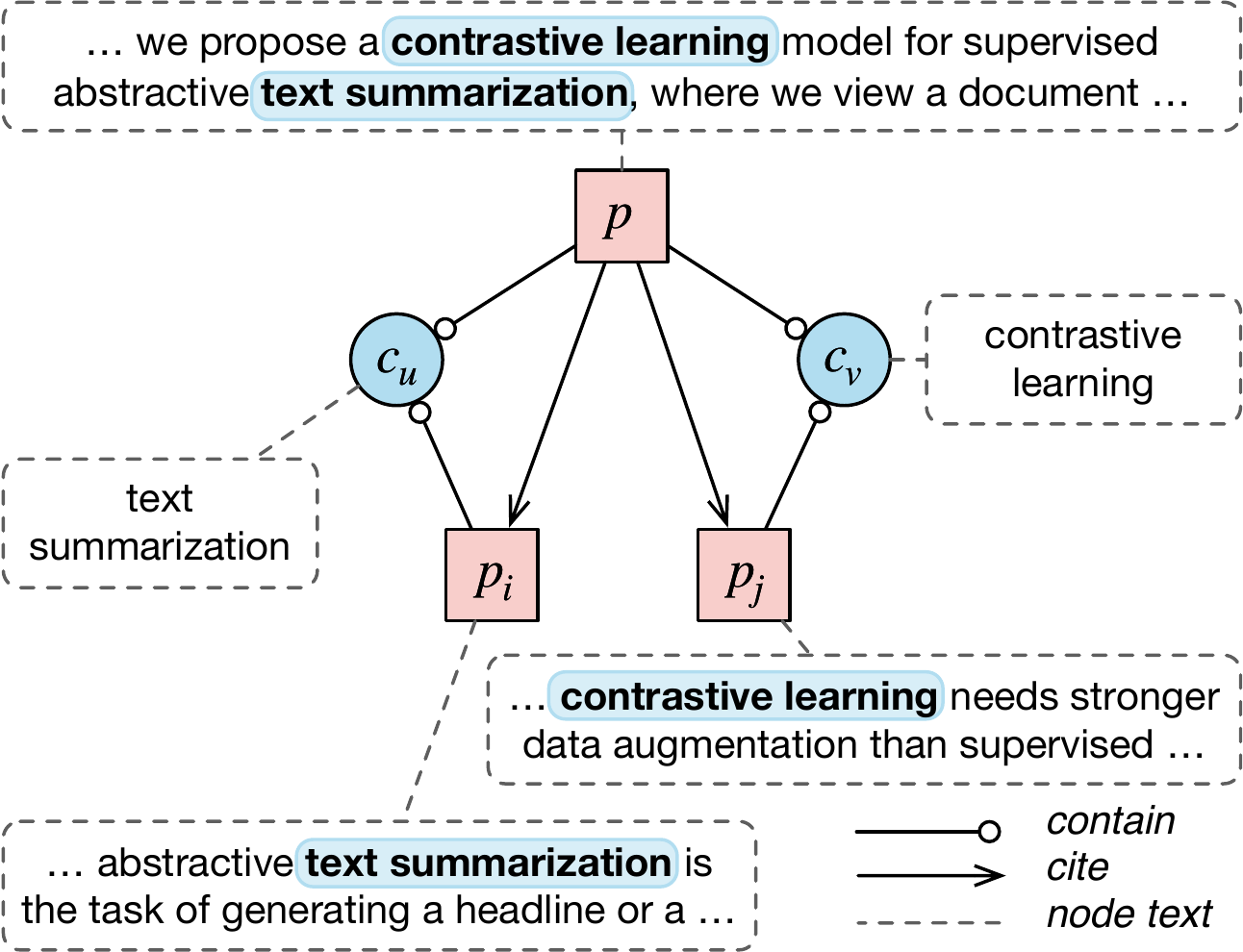}
    \caption{A quintuple with its text attributes. The dashed line and box represent the texts of paper or concept.}
    \label{fig:quintuple}
\end{figure}

\subsection{Co-occurrence Citation Quintuple}
Assuming that paper $p$ contains concept $c_u$ and $c_v$, $p$ cites paper $p_i$ and $p_j$ ($p_i \neq p_j$). Meanwhile, $p_i$ contains concept $c_u$, and $p_j$ contains concept $c_v$. Then, for papers $p_i$, $p_j$, and $p$, there exist co-occurrence citation relations corresponding to concepts $c_u$ and $c_v$. Formally, let $R_p$ denote the set of reference papers of $p$, and we define the set $Q$ of co-occurrence citation quintuples as:

\begin{equation}
    \begin{split}
    Q = \{(p_i, p_j, c_u, c_v, p) | p_i \in R_p, p_j \in R_p, \\
    c_u \in C_{p_i} \cap C_{p}, c_v \in C_{p_j} \cap C_{p}, c_u \neq c_v\},
    \end{split}
\end{equation}
where $p$ is called target paper, $p_i$ and $p_j$ are called reference papers. In practice, we bind sentences that mention related concepts to the quintuples, illustrating how an idea existing in $p$ comes up. Figure~\ref{fig:quintuple} shows an example of such quintuple, which consists of two concepts \textit{text summarization} and \textit{contrastive learning}. In the training process, we use the corresponding texts of $p_i, p_j, c_u$, and $c_v$ as input, and our model is expected to generate the idea sentence in $p$, which usually appears in the paper abstract or introduction section.

\begin{figure*}[htb]
    \centering
    \includegraphics[width=1.0\linewidth]{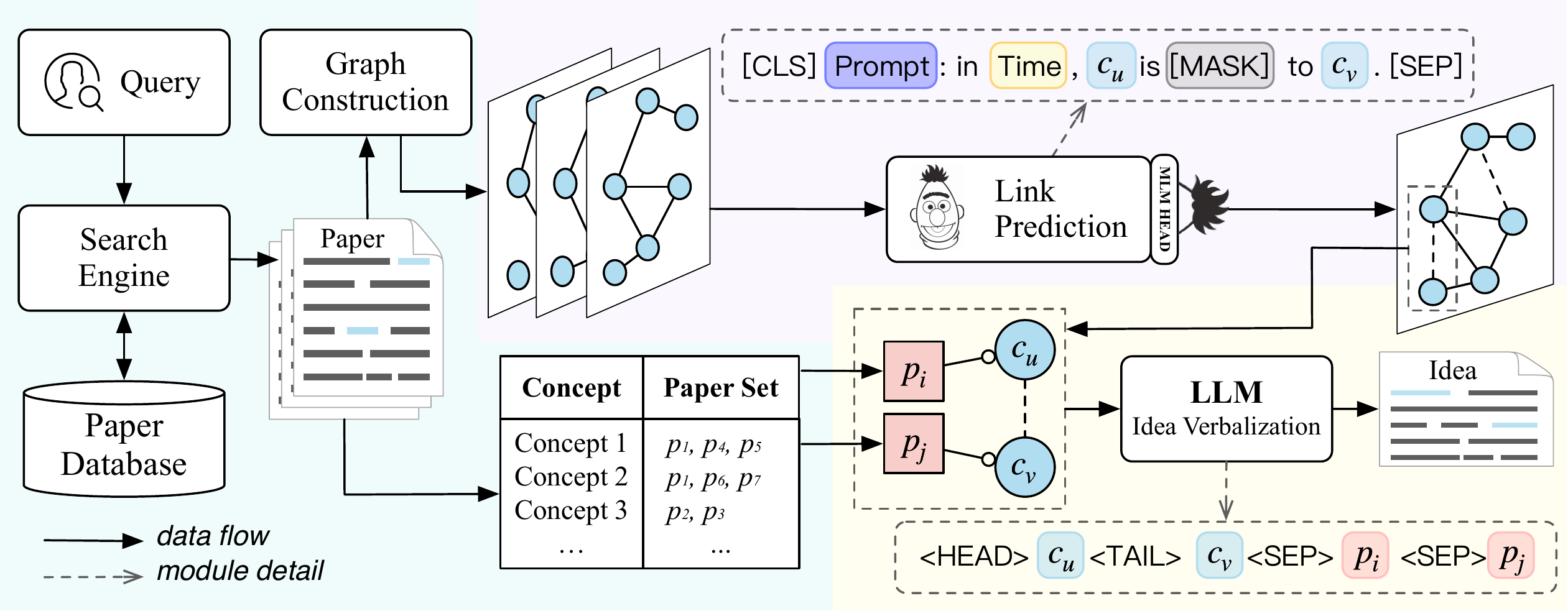}
    \caption{Overview of our research assistant system. The system starts by receiving the user's query and ends with verbalizing an idea. The left part shows the data retrieval and graph construction module. The upper right part is the temporal link prediction module. The lower right part is the idea verbalization module.}
    \label{fig:overview}
\end{figure*}

\section{Datasets and Technical Details}

\subsection{Datasets}\label{sec:data}
Our work relies on a daily updated database containing more than 220 million academic papers from 19 disciplines published between 1800 and 2023. The database also stores nearly 800K concept entities with descriptions. See Appendix~\ref{disofpaper} for the number of papers in each discipline.

To train our model for temporal link prediction, we first collect 240 essential and common queries from 19 disciplines and one special topic (COVID-19). Then, we enter these queries into the paper database to fetch the most relevant papers between 2000 and 2021 with Elasticsearch, a modern text retrieval engine that stores and retrieves papers. Afterward, we use information extraction tools including AutoPhrase~\cite{shang2018automated} to identify concepts. Only high-quality concepts that appear in our database will be preserved. Finally, we construct 240 evolving concept co-occurrence graphs, each containing 22 snapshots according to the co-occurrence relationship. The statistics of the concept co-occurrence graphs are provided in Appendix~\ref{statecc}.

Besides, we construct and release a dataset of co-occurrence citation quintuples, which is used to train text generation model for idea verbalization. We select nearly 9.5M highly-cited papers (500K per discipline) and their corresponding references (19.7M) to construct quintuples. The process of identifying and processing concepts is similar to constructing the concept co-occurrence graph. Heuristic rules are adopted to filter redundant and noisy sentences, further improving the quality of the quintuples used for idea generation. The statistics and more details of co-occurrence citation quintuples can be found in Appendix~\ref{statsofquint}, \ref{filter}, and \ref{officialversion}.

\subsection{Framework Overview}
The framework of our system in the production environment is illustrated in Figure~\ref{fig:overview}. It starts by receiving the user's query and retrieving the most relevant papers from database to construct an evolving concept co-occurrence graph in a real-time way. Meanwhile, the system maintains two dictionaries for storing the mapping relations between papers and concepts. Then, a BERT-based temporal model predicts potential connections of concepts such as $c_u$ and $c_v$, which can be regarded as a new idea. Finally, these connected concepts, as well as their corresponding sentences of papers stored in the above dictionary, are fed to our pretrained model T5 to verbalize an idea. Our system also allows users to select elements they are interested in to form a group of inputs $(p_i, p_j, c_u, c_v)$ for idea verbalization. In the following parts, we will introduce two key components in detail.

\subsection{Temporal Link Prediction}
Our system dynamically constructs a unique evolving concept co-occurrence graph for each query according to the papers retrieved by the search engine. Under the circumstance, a general link prediction model with high transferability is required to predict new connections on different graphs, which means there exists only one set of model parameters. We take advantage of the masked language model (MLM) to tackle the link prediction problem on different graphs and propose a new temporal training method called PLM-LP (See Appendix~\ref{frame-plmlp} for the illustration of PLM-LP).

Given a concept pair $c_u$, $c_v$ and a timestamp $t$, we concatenate these elements and prompt words $pro(c_u, c_v, t)$ to obtain the following input sequence $x_{uv}^t$:
\begin{center}\small
    $x_{uv}^t$ = [CLS] $pro(c_u, c_v, t)$: in $t$, $c_u$ is [MASK] to $c_v$.[SEP],
\end{center}
where $pro$ is a prompt function defined in Equation~\ref{eq:prompt} that generates a description of the given input, [MASK] is the mask token, [CLS] and [SEP] represent the tokens of the beginning and end of the input sequence, respectively. Our model is expected to fill the mask token with a relation token, i.e., $``related"$ and $``unrelated"$, which are taken as the true label to indicate whether the two concepts are connected. Considering that edges in the evolving concept co-occurrence graph do not disappear, we add prompts according to this feature. If there was an edge between $c_u$ and $c_v$ before time $t$, the $pro(\cdot)$ returns the word $``Existing"$, otherwise it returns $``Unknown"$:

\begin{equation}\label{eq:prompt}
    pro(c_u, c_v, t)=\left\{
            \begin{aligned}
                &``Existing", \mathcal{A}_{t-1}(c_u, c_v) = 1\\
                &``Unknown", otherwise
            \end{aligned}
        \right.
\end{equation}

In the data preprocessing, positive samples $\mathbb{D}^{+} = \{x_{uv}^t| \mathcal{A}_{t}(c_u, c_v)=1, T_s \leq t \leq T_e\}$ are directly constructed according to the edges of each year. For negative samples $\mathbb{D}^{-}$, since the concept co-occurrence graph is sparse, we cannot simply take any two concepts that do not have a connection each year as negative samples, which is unreasonable and will lead to a sharp increase in the number of negative samples. Actually, we only need to focus on the samples in the most difficult cases. Therefore, given a concept $c_u \in C$ and its $k$-hop neighborhood concepts, we choose concepts that have no connection with $c_u$ in the next $d$ years to construct negative samples. The set of negative samples is shown as follows:

\begin{equation}
    \begin{split}
    \mathbb{D}^{-} = \{x_{uv}^t|c_v \in \mathcal{N}_k(c_u), \mathcal{A}_{t+d}(c_u, c_v)=0, \\
    k \ge 2, T_s \leq t < t + d \leq T_e\},
    \end{split}
\end{equation}
where $\mathcal{N}_k(c_u)$ is the set of concepts at a distance less than or equal to $k$ from $c_u$, i.e., the k-hop neighborhood of $c_u$. It is worth noting that the negative samples are used to construct input text sequences with timestamp $t$ rather than $t+d$, and we do not generate negative samples in the last $d$ timestamps. We fine-tune the parameters and vocabulary embeddings of BERT via predicting the masked token. Formally, we compute the cross-entropy loss:

\begin{center}
\small    
\begin{equation}
    \mathcal{L}=-\sum_{d \in \mathbb{D}^{+} \cup \mathbb{D}^{-}}1_{[MASK]=y_d}\log P([MASK]=y_d|x^{t}_{uv}),
\end{equation}
\end{center}
where $y_d \in \{``related", ``unrelated"\}$ is the label of the sample. It should be mentioned that KG-BERT~\cite{yao2019kg} and LP-BERT~\cite{da2022multi} are similar to PLM-LP, but the settings they adopt are not applicable to the training of temporal data. Nevertheless, the PLM in our method can be replaced by other models.

\subsection{Idea Verbalization}
In our public beta system, we employ T5~\cite{raffel2020exploring}, a large pretrained sequence-to-sequence model for idea verbalization. We select 2M highly-cited papers for unsupervised denoising training with the language model loss:
\begin{equation}\label{denoising_loss}
    \mathcal{L}_{lm} = \mathbb{E}_{p}[-\log P(p|\tilde{p};\theta)],
\end{equation}
where $\tilde{p}$ represent the corrupted sentence of paper $p$. In the process of fine-tuning, given a co-occurrence citation quintuple $q=(p_i, p_j, c_u, c_v, p)$, we first concatenate $p_i, p_j, c_u$, and $c_v$ to a sequence $Seq(q)$, using $\langle$HEAD$\rangle$, $\langle$TAIL$\rangle$, $\langle$SEP$\rangle$ to denote the head, tail of a concept pair, and the separator, respectively, which is shown as follows:

\begin{center}\small
$Seq(q)$ = $\langle$HEAD$\rangle$ $c_u$ $\langle$TAIL$\rangle$ $c_v$ $\langle$SEP$\rangle$ $p_i$ $\langle$SEP$\rangle$ $p_j$.
\end{center}

We fine-tune the T5 model to find the optimal parameters $\theta^{*}$ to encode the input sequence and verbalize it into an idea sequence, i.e., the item $p$ in the quintuple. For this purpose, we use the maximum likelihood estimation objective:

\begin{equation}
    \theta^{*} = \mathop{\arg\max}\limits_{\theta}\prod_{q}P(p|Seq(q);\theta).
\end{equation}

During the inference process (production environment), we use the predicted connection of concepts $c_u$, $c_v$, and their corresponding sentences of papers $p_i, p_j$ to construct the input sequence, which is encoded by our fine-tuned T5 to generate an idea sequence. Note that the idea verbalization model is also flexible in our framework, and it can be substituted by alternatives such as GPT\cite{radford2018improving} with another configuration of fine-tuning. We will also provide premium subscribers with GPT-3.5 after the official release of our system.

\begin{table*}[htb]
    \small
	\begin{center}
        \setlength\tabcolsep{12pt}
		\begin{tabular}{|c|c|ccc|ccc|}
			\hline
			\multirow{2}{*}{\textbf{Method}} & \multirow{2}{*}{\textbf{Accuracy}} & \multicolumn{3}{c|}{\textbf{All Edges in 2021}} &  \multicolumn{3}{c|}{\textbf{New Edges in 2021}}  \\
			\cline{3-8}
			 &  & \textbf{Precision} & \textbf{Recall} & \textbf{F1} & \textbf{Precision} & \textbf{Recall} & \textbf{F1} \\
			\hline
			SEMNET & 0.478 & 0.099 & 0.519 & 0.146 & 0.007 & 0.552 & 0.013 \\
			GCN-GAN & 0.975 & \textbf{1.000} & 0.860 & 0.924 & N/A & 0 & N/A \\
			EvolveGCN & \textbf{0.995} & \textbf{1.000} & 0.970 & \textbf{0.985} & N/A & 0 & N/A \\ \hline
			PLM-LP w/o \textit{pro}. & 0.648 & 0.586 & 0.948 & 0.646 & 0.467 & 0.947 & 0.474 \\
			PLM-LP \textit{ind}. & 0.742 & 0.704 & 0.986 & 0.748 & 0.188 & 0.910 & 0.195 \\
			PLM-LP & 0.735 & 0.970 & \textbf{0.998} & 0.981 & \textbf{0.540} & \textbf{0.988} & \textbf{0.560} \\
			\hline
		\end{tabular}
	\end{center}
	\caption{Average results of link prediction on different disciplines. The best results are boldfaced. N/A means all cases have been predicted to be negative.}
	\label{tab:ave_subject_result}
\end{table*}

\section{Evaluation}
\subsection{Analysis of Temporal Link Prediction}
\subsubsection{Results of Link Prediction in 2021}
PLM-LP is compared with 3 temporal model SEMNET~\cite{Krenn2020PredictingRT}, GCN-GAN~\cite{lei2019gcn}, and EvolveGCN~\cite{pareja2020evolvegcn}, which are suitable for concept co-occurrence graph. SEMNET analyzes graph characteristics to recognize potential new edges with an MLP module. GCN-GAN and EvolveGCN utilize GCN and LSTM to model the structural and temporal information of a graph. In the experiment, their performance is evaluated on our constructed 240 concept co-occurrence graphs, where the last snapshot (the year 2021) is used as the test set. We report the accuracy of the adjacent matrix, precision, recall, and F1 score of all edges and new edges existing in the graph of 2021. New edges do not exist in the past snapshots and only come out in 2021. 

Note that PLM-LP is trained with a single set of model parameters on these 240 graphs and then applied to different graphs for the test procedure. The hyper-parameters $k$ and $d$ in PLM-LP are set to 2 and 5, respectively. Apart from our proposed PLM-LP, we also introduce two variants. PLM-LP w/o \textit{pro.} removes the prompt words $pro(c_u,c_v,t)$. PLM-LP \textit{ind}. is trained with independent parameters on different graphs. Results of these models in 20 disciplines/topics are provided in Appendix~\ref{cmp-lp-all}. The average results are shown in Table~\ref{tab:ave_subject_result}. It can be observed that all these models are capable of identifying most edges existing in 2021, but the GCN-GAN and EvolveGCN gets undesirable performance to find new edges in 2021. Many cases have been predicted to be unconnected. We believe this is because most graphs are sparse, leading to overfitting. In our scenario, detecting new edges is more important than improving the accuracy of the adjacency matrix. Our proposed method can tackle the issue to a certain extent. As to the variants, it is difficult for PLM-LP w/o \textit{pro.} to correctly predict all edges in 2021 due to the absence of prompt words. PLM-LP \textit{ind}. is also inferior to PLM-LP, indicating that PLM can learn interdisciplinary knowledge with a set of training parameters.

\subsubsection{Human Assessment of Link Prediction in the Future}
We use all graph snapshots, including the year 2021, for training to mine potential connections that may appear in the future. Similarly, we select the top 20 pairs of concepts for each query. See Appendix~\ref{plmlppre} for the potential connections of different disciplines. We invited more than 10 experts from the field of computer science and geo-science (geology and geography) to evaluate the predicted results in their corresponding domains. The assessment is based on the experience of experts. The results are shown in Table~\ref{tab:human_link_prediction}. As expected, at least a third of the potential concept pairs predicted by the system are reasonable in the three disciplines, indicating that PLM-LP is able to explore new concepts across disciplines. We also test random pairs on geo-science, and there are no more than 10\% of reasonable pairs.

\begin{table}[htb]
\small
\centering
\setlength\tabcolsep{17pt}
\begin{tabular}{|l|c|}
\hline
\multicolumn{1}{|c|}{\textbf{Disciplines}} & \textbf{\begin{tabular}[c]{@{}c@{}}Percentage (\%) of \\ Reasonable Pairs\end{tabular}} \\ \hline
Computer Science                           & 52.1                                    \\ \hline
Geology                                    & 48.8                                    \\ \hline
Geography                                  & 34.2                                    \\ \hline
\end{tabular}
\caption{Percentage (\%) of reasonable concept pairs based on human assessment.}
\label{tab:human_link_prediction}
\end{table}

\subsection{Analysis of Idea Verbalization}
\subsubsection{Benchmark Results}
We release the co-occurrence citation quintuples for idea verbalization, which can be used as a benchmark for natural language generation. Our public beta system adopts PLM such as T5 and BART as the generation models that are fine-tuned on the quintuples. We also apply unsupervised denoising training on T5 with highly-cited papers, which makes the PLM itself learn more academic knowledge. All training and inference processes are carried out on NVIDIA GeForce RTX 3090. In the fine-tuning stage, we employ Adam as the optimizer with 0.01 weight decay. The learning rate is set to 1e-4. For the inference, the beam size is set to 4. Similar to previous text generation work~\cite{fan2018hierarchical, wang2019paperrobot}, we use BLEU~\cite{papineni2002bleu}, METEOR~\cite{banerjee2005meteor}, and ROUGE\_L~\cite{lin2004rouge} to measure the fluency and topic relevance of the generated ideas. Table~\ref{tab:benchmark_results} gives the benchmark results.

\begin{table}[htb]
\centering
\small
\setlength\tabcolsep{6pt}
\begin{tabular}{|l|ccc|}
\hline
 \textbf{Model}        & \textbf{BLEU} & \textbf{METEOR} & \textbf{ROUGE\_L} \\ \hline
T5-base  & 25.16         & 12.57           & 16.66             \\ 
T5-large & 25.68         & 12.72           & 16.83             \\ 
T5-base \textit{denoise}  & 25.72         & 12.54           & 16.74             \\ 
T5-large \textit{denoise} & 26.94         & 13.19           & 17.35             \\ 
BART-large & 21.87         & 7.93       & 14.72         \\ \hline
\end{tabular}
\caption{Benchmark results with different PLMs.}
\label{tab:benchmark_results}
\end{table}

In fact, it is challenging to evaluate long text~\cite{liu2016not,li2016persona}, let alone idea verbalization, which may contain new opinions, insights, and methods. Additionally, the new content in the verbalized idea is likely to differ from the target paper in quintuples. Thus, we conduct the following experiments.

\subsubsection{Turing Test}

\begin{table*}[htb]
\setlength\tabcolsep{15pt}
\small
\centering
\begin{tabular}{|l|c|c|c|cc|}
\hline
\multicolumn{1}{|c|}{\multirow{2}{*}{\textbf{Disciplines}}}                     & \multirow{2}{*}{\textbf{Test ID}} & \multirow{2}{*}{\textbf{\# Cases}} & \multirow{2}{*}{\textbf{\begin{tabular}[c]{@{}c@{}}\# Options \\ per Case\end{tabular}}} & \multicolumn{2}{c|}{\textbf{\# Participant}}                   \\ \cline{5-6} 
\multicolumn{1}{|c|}{}                                                          &                                   &                                    &                                                                                          & \multicolumn{1}{c|}{\textbf{\# Amateur}} & \textbf{\# Expert}  \\ \hline
\multirow{2}{*}{\begin{tabular}[c]{@{}l@{}}Computer  Science\end{tabular}}    & 1.1                               & 50                                 & 2                                                                                        & \multicolumn{1}{c|}{\multirow{2}{*}{10}} & \multirow{2}{*}{30} \\
                                                                                & 1.2                               & 20                                 & 3                                                                                        & \multicolumn{1}{c|}{}                    &                     \\ \hline
\multirow{2}{*}{\begin{tabular}[c]{@{}l@{}}Geography \& Geology\end{tabular}} & 2.1                               & 30                                 & 2                                                                                        & \multicolumn{1}{c|}{\multirow{2}{*}{6}} & \multirow{2}{*}{6} \\
                                                                                & 2.2                               & 20                                 & 3                                                                                        & \multicolumn{1}{c|}{}                    &                     \\ \hline
\multirow{2}{*}{\begin{tabular}[c]{@{}l@{}}Medicine \& COVID-19\end{tabular}} & 3.1                               & 30                                 & 2                                                                                        & \multicolumn{1}{c|}{\multirow{2}{*}{8}} & \multirow{2}{*}{10} \\
                                                                                & 3.2                               & 20                                 & 3                                                                                        & \multicolumn{1}{c|}{}                    &                     \\ \hline
\end{tabular}
\caption{Settings of Turing test.}
\label{tab:turing_setting}
\end{table*}

Similar to previous work~\cite{wang2019paperrobot}, we recruited more domain experts and non-experts in the field of computer science, geo-science (geology and geography), and medicine to conduct the Turing test. Experts include professors, lecturers, postdoctoral researchers, and graduate students (at least two professors per discipline). Participants are asked to read the machine-generated outputs and human-written texts and choose the real human-written text from a set of $N-1$ fake ones. Each participant is given instructions before the test. We also allow participants to use the Internet to retrieve technical terms during the test. For each discipline, there are two different modes of multiple-choice questions, one contains two options per question, and the other contains three options per question. We randomly select 15 questions per test from the question bank for each participant to answer. We conduct six groups of Turing tests, whose experimental settings are shown in Table~\ref{tab:turing_setting}.

\begin{figure}[htb]
    \centering
    \includegraphics[width=1.0\linewidth]{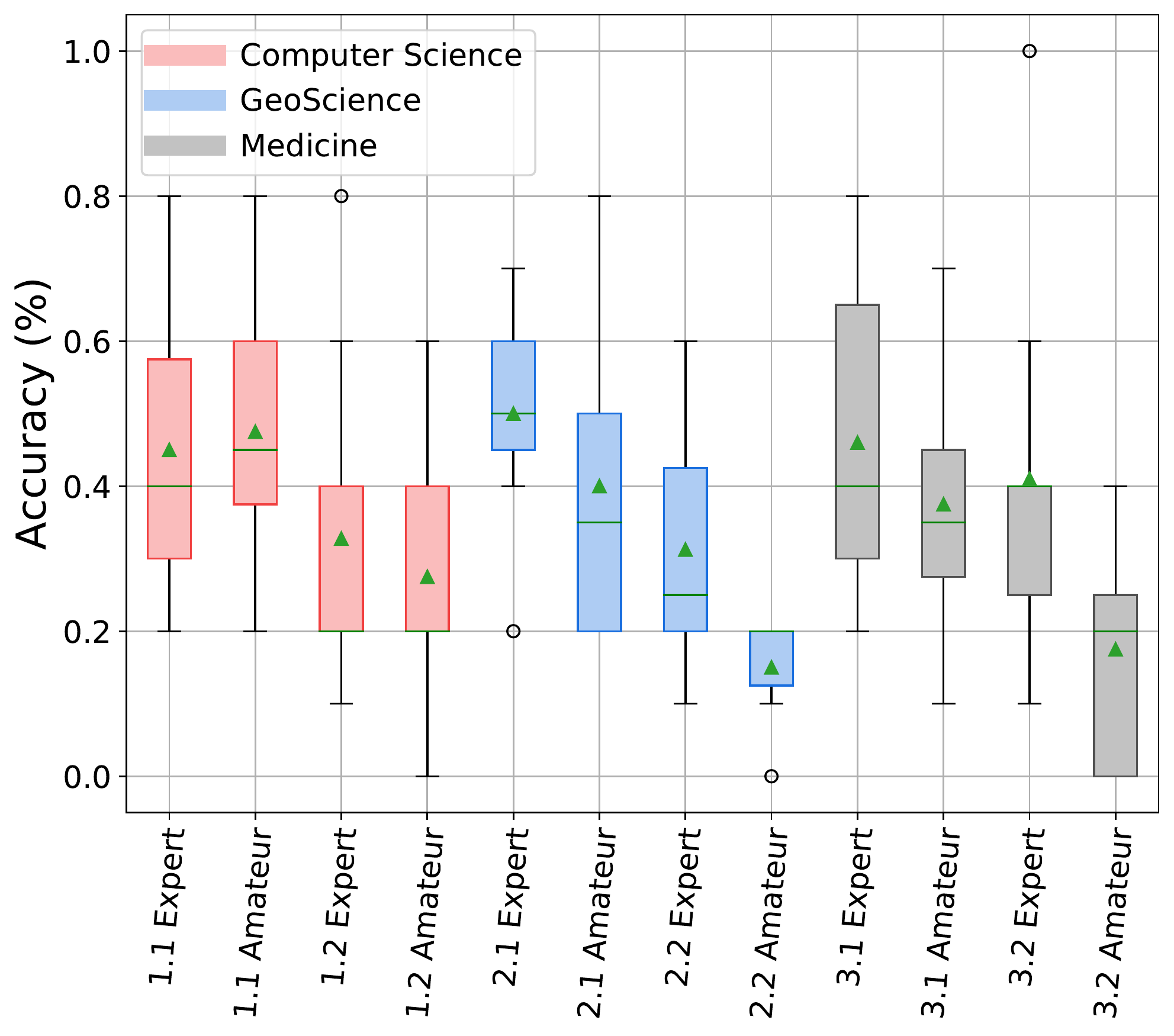}
    \caption{Box plot of Turing test. The green triangle represents mean value, and the green line represents median value. The label of the x-axis is composed of the test ID and participant role.}
    \label{fig:turing_result}
\end{figure}

The results are displayed using a box plot in Figure~\ref{fig:turing_result}. Overall, domain experts are more likely to achieve higher accuracy in these six groups of tests. Also, the results reveal that the accuracy of the 3-options question is lower than 30\%, indicating that it is more difficult for participants to choose the human-written one from 3 options than from 2 options. Moreover, the accuracy of the 2-option questions is close to or even lower than that of random guessing, which means experts can hardly distinguish between human-written sentences and machine-generated sentences, although they tend to analyze texts from the perspective of logic and accuracy. One of the possible reasons is that the verbalized ideas contain more non-professional terms while maintaining fluency and reasonableness, which is more readable than academic papers.

\subsubsection{Relevance \& Plagiarism Analysis}
We calculate the percentage of n-grams in the input sequence which appear in the verbalized idea of test data to analyze how relevant the idea is to the input sequence. Meanwhile, the percentage of n-grams can also be regarded as a plagiarism check. As seen from Table~\ref{tab:plagiarism}, about 40\% of the input 1-grams exist in the output texts, which means the output can combine the knowledge of relevant concepts. Additionally, the percentages of 2 to 5-grams are all lower than 20\%, that is, the verbalized ideas are not simply copied from the input but are paraphrased and fused into new knowledge.

\begin{table}[ht]
\centering
\small
\setlength\tabcolsep{8pt}
\begin{tabular}{|c|c|c|c|c|}
\hline
\textbf{1-gram} & \textbf{2-gram} & \textbf{3-gram} & \textbf{4-gram} & \textbf{5-gram} \\ \hline
40.7            & 19.9      & 13.8             & 11.2              &  9.4           \\ \hline
\end{tabular}
\caption{Percentage (\%) of n-grams in the test input sequence that appear in the verbalized idea.}
\label{tab:plagiarism}
\end{table}

We did not provide the n-gram overlap of the target paper $p$. From our perspective, $p$ is more like an inspiration, which teaches our model a possible way of how two concepts can be combined together, rather than generating sentences exactly like $p$.

\begin{table*}[htb]
\centering
\setlength\tabcolsep{1pt}
\small
\begin{tabular}{|l|c|c|c|}
\hline
\textbf{Discipline} & \textbf{Computer Science} & \textbf{Economics} & \textbf{Political Science}\\ \hline
\textbf{Connection}                                                & \textbf{\begin{tabular}[c]{@{}c@{}}knowledge graph \&\\ contrastive learning\end{tabular}} & \textbf{\begin{tabular}[c]{@{}c@{}}intellectual capital \& \\ income distribution\end{tabular}} & \textbf{\begin{tabular}[c]{@{}c@{}}gender equity \&\\ economic crisis\end{tabular}} \\ \hline
\textbf{\begin{tabular}[c]{@{}l@{}}Verbalized\\ Idea\end{tabular}} & \multicolumn{1}{l|}{\begin{tabular}[c]{@{}l@{}}We present a new approach to \\ \textbf{knowledge graph} mining that\\  leverages ontologies. The key \\ idea is to model the domain \\ knowledge as a \textbf{tree like structure} \\ with nodes and edges connected \\ in a directed or unordered graph. \\ This allows us to efficiently learn\\  from large amounts of \textbf{unlabeled} \\ \textbf{data without having to manually} \\ \textbf{annotate it}. Experiments show \\ that this approach outperforms \\ existing approaches such as tree \\ augmented neural networks and \\ SVM for both text classification \\ and image categorization tasks \\ where they only use small subsets \\ of training examples.\end{tabular}} & \multicolumn{1}{l|}{\begin{tabular}[c]{@{}l@{}}This paper examines the determin-\\ ants of inequality in income and \\ \textbf{wealth distributions}, with particu-\\ lar emphasis on the role played \\ by \textbf{intellectual property rights}. \\ It is shown that there are two main \\ sources of inequality: unequal \\ \textbf{income distributions} due to \\ ownership of patents and trade \\ secrets, and unequal taxation of \\ these same properties. The first \\ source arises from the fact that most \\ \textbf{patents are owned by individuals} \\ who do not have control over their \\ own economic activities. In contrast, \\ the second source derives from \\ the existence of \textbf{private property} \\ \textbf{rights} which make it possible for an \\ individual to be rich without having to \\ pay taxes on his or her own income.\\ \end{tabular}} & \multicolumn{1}{l|}{\begin{tabular}[c]{@{}l@{}}During the global financial and \\ economic crisis, \textbf{women’s employment} \\ \textbf{opportunities declined sharply}. These \\ trends are likely to continue during \\ the next few years as more women \\ enter the labor force. From our \\ perspective, there will be an increasing \\ number of women entering the work-\\ force at \textbf{lower levels of education} than \\ men. This trend is expected to continue \\ in the coming years as female particip-\\ ation in the labour force continues to \\ increase. The current \textbf{political and} \\ \textbf{economic climate} may make it difficult \\ for women to access higher level \\ education because of the challenges \\ presented by the \textbf{gender pay gap} and \\ the macroeconomic crisis that has \\ gripped much of the \textbf{developing world} \\ since 2007.\end{tabular}} \\ \hline
\end{tabular}
\caption{Case study in computer science, economics, and political science.}
\label{tab:case_study}
\end{table*}

\subsection{Case Study}
In Appendix~\ref{expturing}, we provide a page of examples of input sequences, human-written texts, and verbalized ideas according to our test dataset of quintuples. To simulate the real situation, we randomly select cases including new connections PLM-LP predicts, which do not appear in our quintuple dataset. It is worth noting that we only take these two concepts as input and do not enter their corresponding sentences to avoid the impact of potential plagiarism. 

Table~\ref{tab:case_study} shows three verbalized ideas. For the first case, we can see our system integrates the critical characteristic of contrastive learning that requires no labels into the task of knowledge graph mining. However, it includes untested experimental results due to the denoising training from numerous papers (especially from paper abstracts and introduction section), and we remove them with heuristic rules in the production environment. As to the second case, the verbalized idea mentions that intellectual capital, such as intellectual property rights, is closely related to income distribution. In the last case, our system believes that a gender pay gap exists in developing countries, which is more obvious during the economic crisis. These cases show that our system can well predict and verbalize ideas, and the generated results align with human intuition and value. Nevertheless, more details are required in natural and exact sciences.

\section{Related Work}
\subsection{Graph Technology for Academic Discovery}
There are a few graph technical methods to help researchers find new ideas. SEMNET~\cite{Krenn2020PredictingRT} predicts research trends with an MLP in the field of quantum physics via constructing such co-occurrence graphs. \citeauthor{sarica2021idea} proposes a technology graph to stimulate idea generation in engineering design, which aims to discover new concepts in the white space surrounding a focal design domain according to the semantic distance. Besides, InfraNodus~\cite{paranyushkin2019infranodus}, a commercial tool for people in different industries, generates insights by detecting structural gaps in a text network, which is similar to mind maps.

\subsection{Text Generation}
Pretrained language models, including T5~\cite{raffel2020exploring}, BART~\cite{lewis2020bart}, and GPT~\cite{radford2018improving} have become the mainstream modules of text generation since they contain billions of parameters and use a large number of corpus for training to achieve good performance. As to text generation for academic research, existing models can only be applied to a few disciplines with much fewer papers than ours. They also require a lot of resources to construct knowledge bases. For instance, PaperRobot~\cite{wang2019paperrobot} adopts external domain knowledge graphs to incrementally generate titles, abstracts, and conclusions of a paper. DRAW~\cite{liu2021train} consists of \textit{reader}, \textit{writer}, and \textit{reviewer} components to generate scientific texts. ChatGPT~\cite{chatgpt2022} generates human-level texts with proximal policy optimization, but it requires professional prompts to discover new ideas. Galactica~\cite{taylor2022galactica} is a large language model for science, which can be combined with our link prediction model to enhance its explainability for idea verbalization.

\section{Conclusion}
We model the emergence of a new idea as two sequential processes: temporal link prediction for exploration and text generation for verbalization. To achieve the objectives, we first construct and release two datasets with new data structures, including evolving concept co-occurrence graph and co-occurrence citation quintuple. Then, we devise a new temporal link prediction method based on the masked language model, which can be applied to various evolving concept co-occurrence graphs of different disciplines. Finally, we fine-tune a PLM to verbalize ideas using the released quintuples. The pipeline has been integrated into a system free for researchers to obtain inspiration. From the experiments and the feedback of users, our system can provide useful information for idea discovery. In the future, we will release an academic oriented language model with the paradigm of prompt learning and instruction tuning to tackle both link prediction and text generation.

\section*{Limitations}
Based on internal review and user feedback, we summarized the following limitations to improve and iteratively update our system and framework in the future.
\\ \textbf{Problem Modeling}: New concepts appear yearly in the real world, but the current system cannot generate new concepts. Generally, the emergence of new concepts often comes from the fusion of mature technologies. Thus, we model the idea exploration as link prediction. Note that it is not the only pathway to brew new ideas, but we have verified the effectiveness and rationality of this approach in the experiments. In addition, PLM can be taken as an implicit knowledge graph~\cite{petroni2019language,wang2020language}, which is capable of tackling uncovered concepts in the evolving concept graphs. We will continue exploring the potential of PLM in knowledge discovery and innovation.
\\ \textbf{Logic, Correctness, and Concreteness}: Although the verbalized ideas can deceive many experts, they may still lack logic, correctness, and details, especially in natural and exact sciences. It is also a challenge for natural language generation. We plan to use more academic corpus and introduce constraint~\cite{zhang2020pointer} to alleviate such problems.
\\ \textbf{Temporal Information}: In PLM-LP, we simply take the year information as a token in the input sequence. We conduct additional experiments to show that the temporal information is not sensitive to PLM-LP, which can be attributed to the negative sampling and the nature of the strictly evolving network.
\\ \textbf{Two Birds One Stone}: The current system employs two different PLMs for link prediction and idea verbalization, respectively. The development of prompt learning~\cite{liu2021pre} reveals that most NLP problems can be regarded as generation problems. In the future, we will introduce new training settings using a single PLM to address link prediction and idea verbalization simultaneously.

\section*{Ethics Statement}
The datasets used in our research are collected through open-source approaches. The whole process is conducted legally, following ethical requirements. As for the Turing Test in our study, all participants are well informed about the purpose of experiments and the usage of test data, and we would not leak out or invade their privacy.

We see opportunities for researchers to apply the system to idea discovery, especially for interdisciplinary jobs. We encourage users to explore different combinations of subjects with the help of our system, making the most of its knowledge storage and thus maximizing the exploration ability of the system.

The main focus of the system is to provide a possible direction for future research, but the effect of human researchers will never be neglected. 

The massive data from various disciplines behind the system makes it capable of viewing the knowledge of an area in a multi-dimensional perspective and thus helps promote the development of novel interdisciplinary. However, considering the risks of misinformation generated by NLP tools, the verbalization only contains possible insights into new ideas. Researchers must thoroughly consider whether an idea is feasible or leads to adverse societal effects.

\section*{Acknowledgements}
We would like to express our deepest gratitude to the scientists involved in the Deep-time Digital Earth program, whose contributions have been incredibly valuable. Additionally, we extend our thanks to Zhongmou He, Jia Guo, Zijun Di, Shengling Zhu, Yanpeng Li, Qi Li, Jiaxin Ding and Tao Shi from the IIOT Research Center at Shanghai Jiao Tong University for their unwavering support during the development of our system. This work was supported by NSF China (No.42050105, 62020106005, 62061146002, 61960206002), Shanghai Pilot Program for Basic Research - Shanghai Jiao Tong University.

\bibliography{anthology,custom}
\bibliographystyle{acl_natbib}

\clearpage

\appendix\label{sec:appendix}

\section{Distribution of Papers}\label{disofpaper}
\begin{figure}[tbh]
    \centering
    \includegraphics[width=1.0\linewidth]{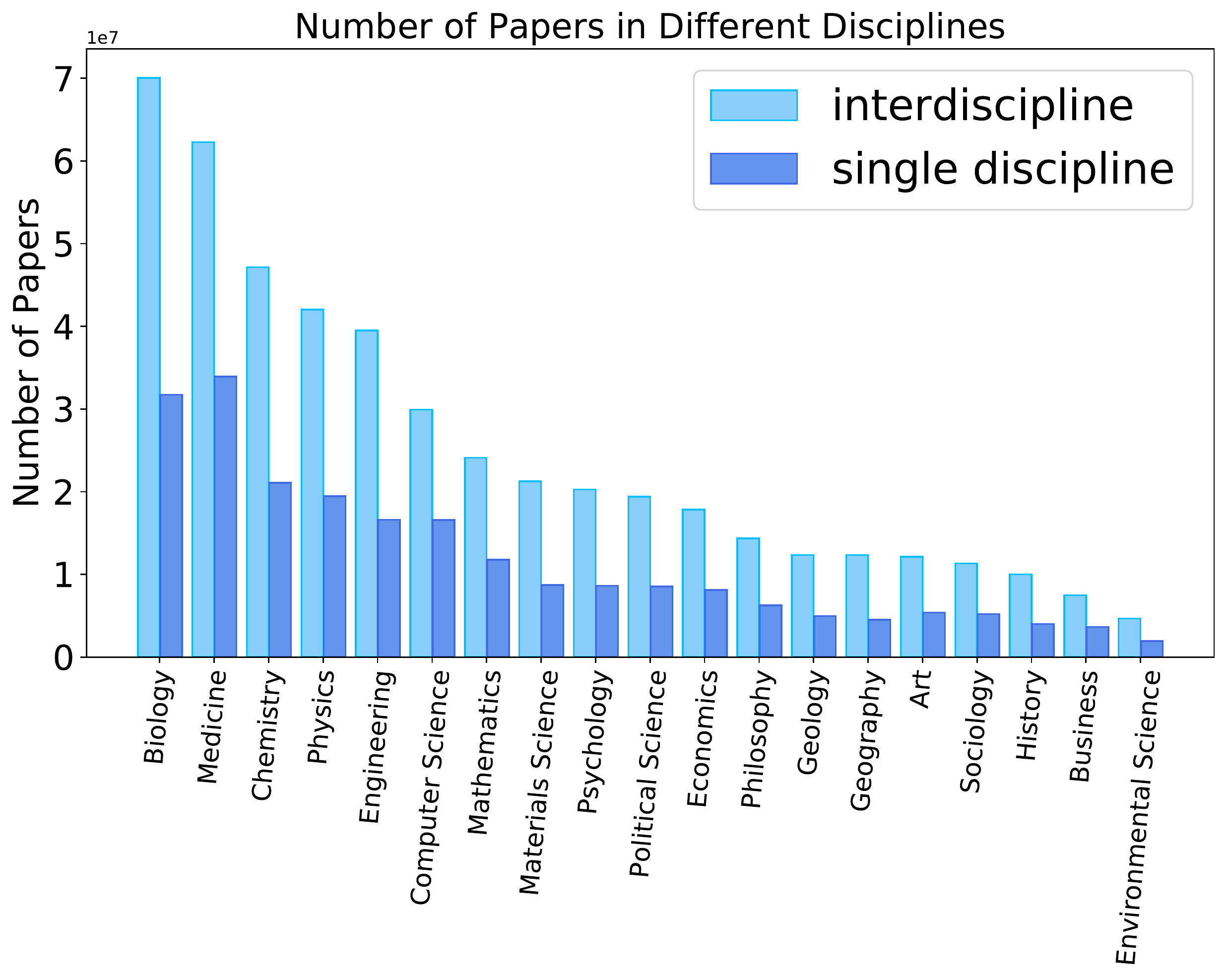}
    \caption{Number of papers in different disciplines.}
    \label{fig:disciplines}
\end{figure}

We are an academic service provider with a sufficient number of high-quality literature data sources (including publications and preprints). These sources are reliable and maintained by a team of professional engineers, ensuring the accuracy and persuasiveness of idea-discovery results. Our database contains more than 220 million academic papers from 19 disciplines between 1800 and 2023 and nearly 800K concept entities with corresponding descriptions. Figure~\ref{fig:disciplines} shows the number of papers in each discipline. Note that there are a large number of interdisciplinary papers. Our system will retrieve relevant papers from this database according to the queries and guide users to discover new ideas.

\begin{table}[h]
\center
\begin{tabular}{|l|r|}
\hline
\textbf{Item}          & \textbf{Count}          \\ \hline
Target Paper           & 9,500,000               \\
Reference Paper        & 19,790,411              \\
Citation Threshold     & 2                      \\
Concept                & 18,347                  \\
Quintuple              & 652,809                 \\ \hline
High-quality Quintuple & 92,313                   \\
Train                  & 73,852                   \\
Valid                  & 9,230                    \\
Test                   & 9,231                    \\ \hline
\end{tabular}
\caption{Statistics of co-occurrence citation quintuples.}
\label{table:quintuple_stats}
\end{table}

\begin{figure}[tbh]
    \centering
    \includegraphics[width=1.0\linewidth]{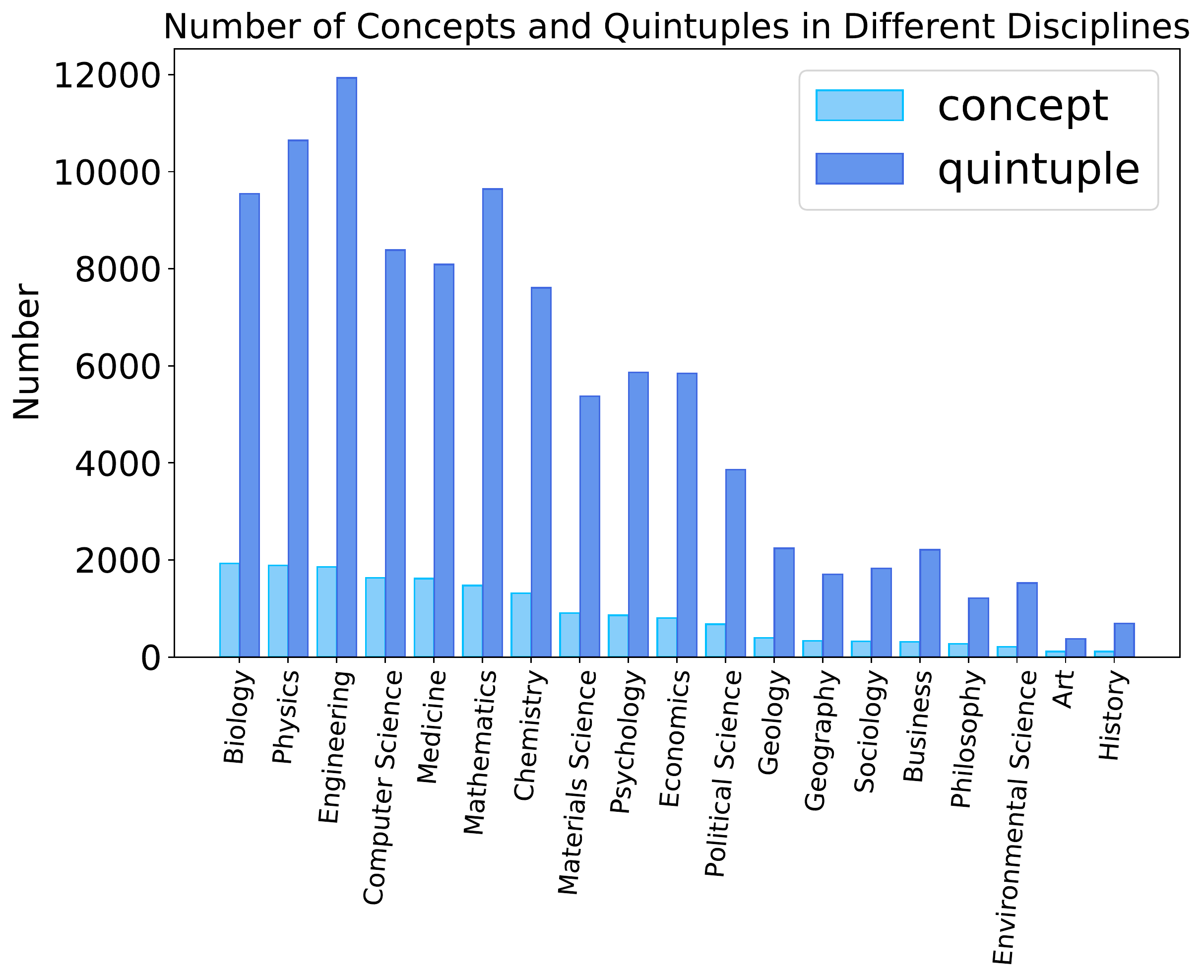}
    \caption{Number of concepts and quintuples in different disciplines.}
    \label{fig:concepts}
\end{figure}

\section{Statistics of Quintuples}\label{statsofquint}
Table~\ref{table:quintuple_stats} shows the statistics of co-occurrence citation quintuples, which originate from 9.5M target papers and 19.8M reference papers. Their citations are greater than or equal to 2. In the data preprocessing, when a paper contains multiple sentences corresponding to a concept, we randomly picked up one sentence to construct a quintuple. We finally obtain 92,313 high-quality instances (73,852 for training, 9,230 for validation, and 9231 for testing) after applying a filter mechanism (Appendix~\ref{filter}). The distribution of the quintuples and their corresponding concepts are shown in Figure~\ref{fig:concepts}. We can see that the numbers of quintuples and concepts of natural science are far more than those of social science, which can be attributed to the paper distribution and citation. In the future, we will lower the citation threshold to get more quintuples of social science.

\section{Pipeline of Quintuple Construction}\label{filter}

\begin{figure*}[tbh]
    \centering
    \includegraphics[width=0.9\linewidth]{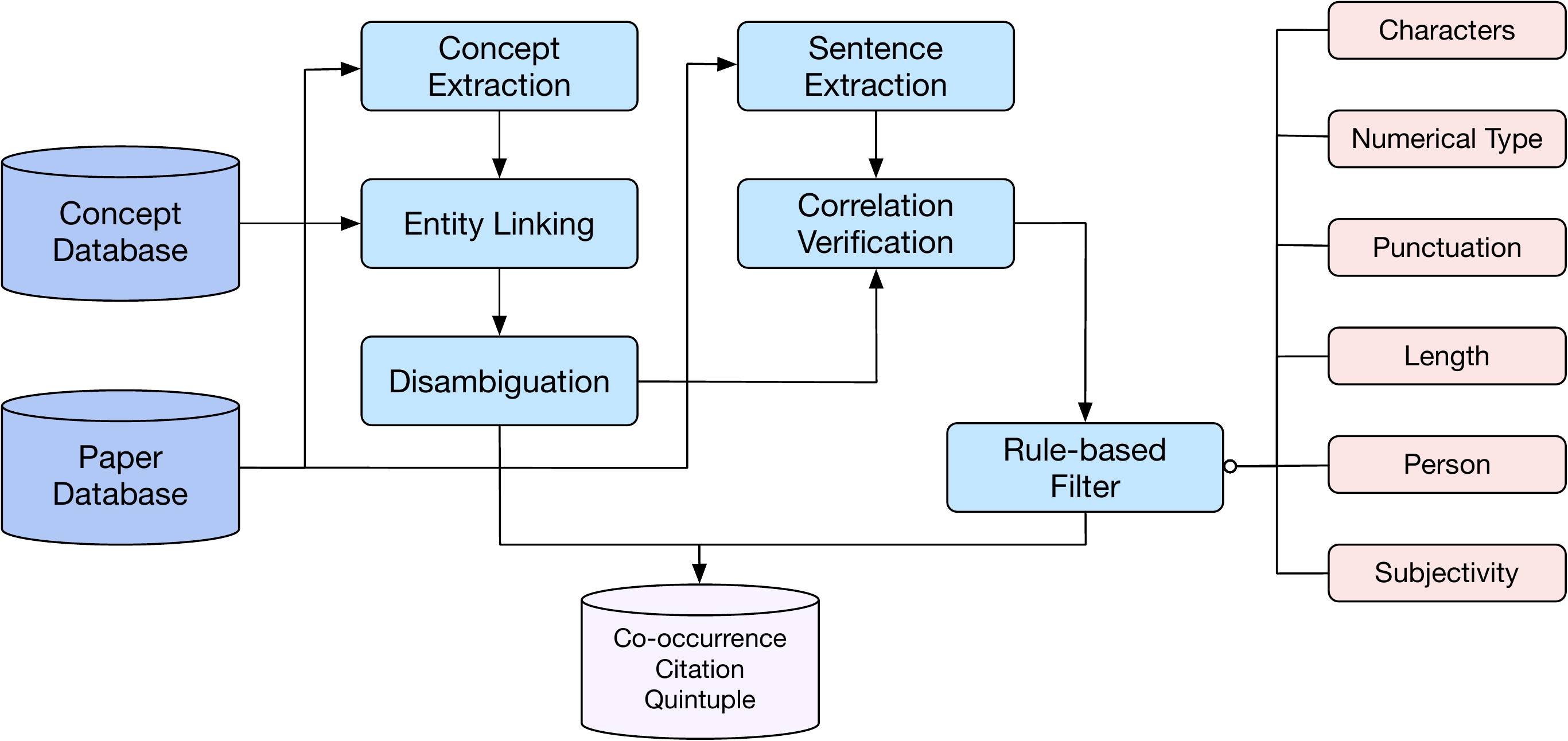}
    \caption{Pipeline of constructing quintuples.}
    \label{fig:quintuple_pipeline}
\end{figure*}

Figure~\ref{fig:quintuple_pipeline} illustrates the pipeline of constructing quintuples. We select nearly 9.5M highly cited papers (500K per discipline) and their corresponding references (19.7M) to construct quintuples. We employ AutoPhrase~\cite{shang2018automated}, an information extraction tool to identify concepts. We execute the process of entity linking and alignment to disambiguate duplicate entities and remove low-quality concepts. Then, we retrieve corresponding sentences of papers that mention these concepts. Relevant sentences will be preserved. Additionally, we apply a rule-based filter to our retrieved contents, where sentences including experimental details, acknowledgments, and sentences with a large number of numerical conclusions, etc., are removed. Finally, we obtain 92,313 quintuples.

\section{Framework of PLM-LP}\label{frame-plmlp}
\begin{figure*}[tbh]
    \centering
    \includegraphics[width=1.0\linewidth]{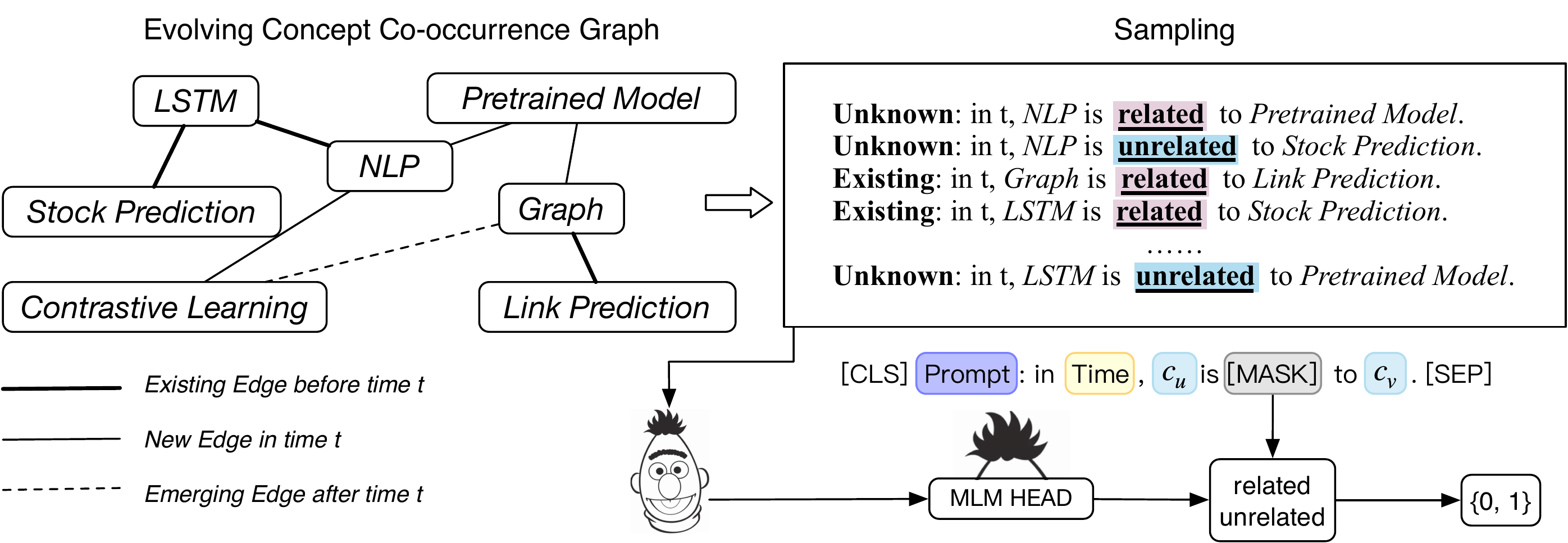}
    \caption{Framework of our proposed temporal link prediction model PLM-LP.}
    \label{fig:PLM-LP}
\end{figure*}

The framework of the temporal link prediction model PLM-LP is illustrated in Figure~\ref{fig:PLM-LP}. We first generate positive and negative samples according to the structure of evolving concept co-occurrence graphs. Note that we add prompt ($``Existing"$ and $``Unknown"$) as the prefix of a sentence. The PLM aims to fill the mask token with a relation token, i.e., $``related"$ and $``unrelated"$. We use a masked language model BERT as the base PLM. We fine-tune the parameters and vocabulary embeddings of BERT via minimizing cross-entropy loss. Note that we simply take the year information as a token in the input sequence. We conduct experiments to show that the temporal information is not sensitive to PLM-LP. In the future, we will design a novel temporal prompt to capture more temporal information.

\section{Examples of Turing Test}\label{expturing}
Table \ref{table:quintuples_examples} shows the examples (2-option questions) used in the Turing Test. All texts presented in the questions originate from the same quintuple, where the human-written text is extracted from the target paper, and the machine-generated text is the idea verbalized by our T5 model according to the concept pair and their corresponding texts. With randomness, repeating the verbalizing process can generate different outputs, which is helpful in preparing questions that need multiple machine-generated texts. From these examples, we can see that machine-verbalized ideas can easily deceive domain experts.

\section{Screenshot of User Interface}\label{interface}
\begin{figure*}[tbh]
    \centering
    \includegraphics[width=1.0\linewidth]{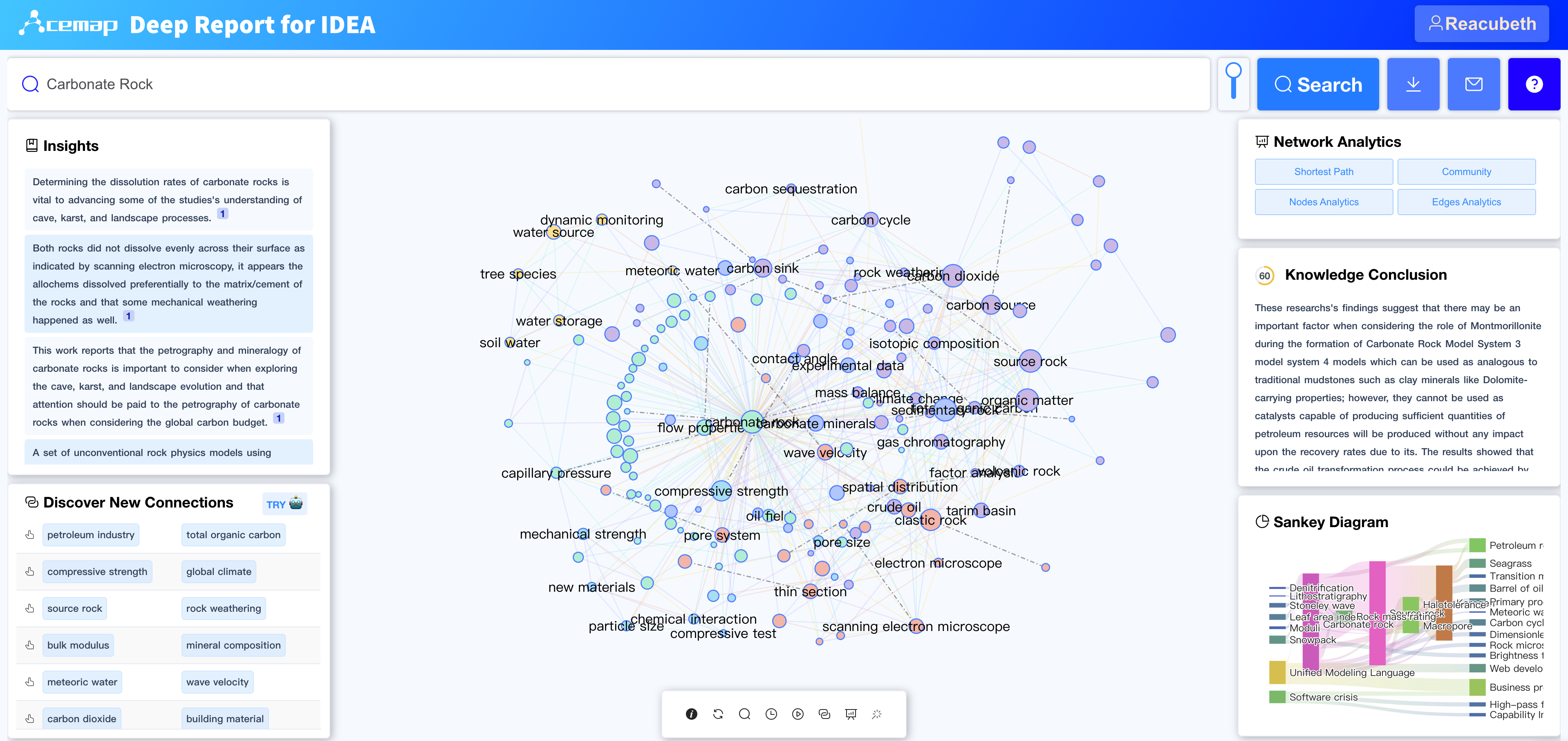}
    \caption{Screenshot of user interface: overview (public beta version).}
    \label{fig:screenshot1}
\end{figure*}

\begin{figure*}[tbh]
    \centering
    \includegraphics[width=1.0\linewidth]{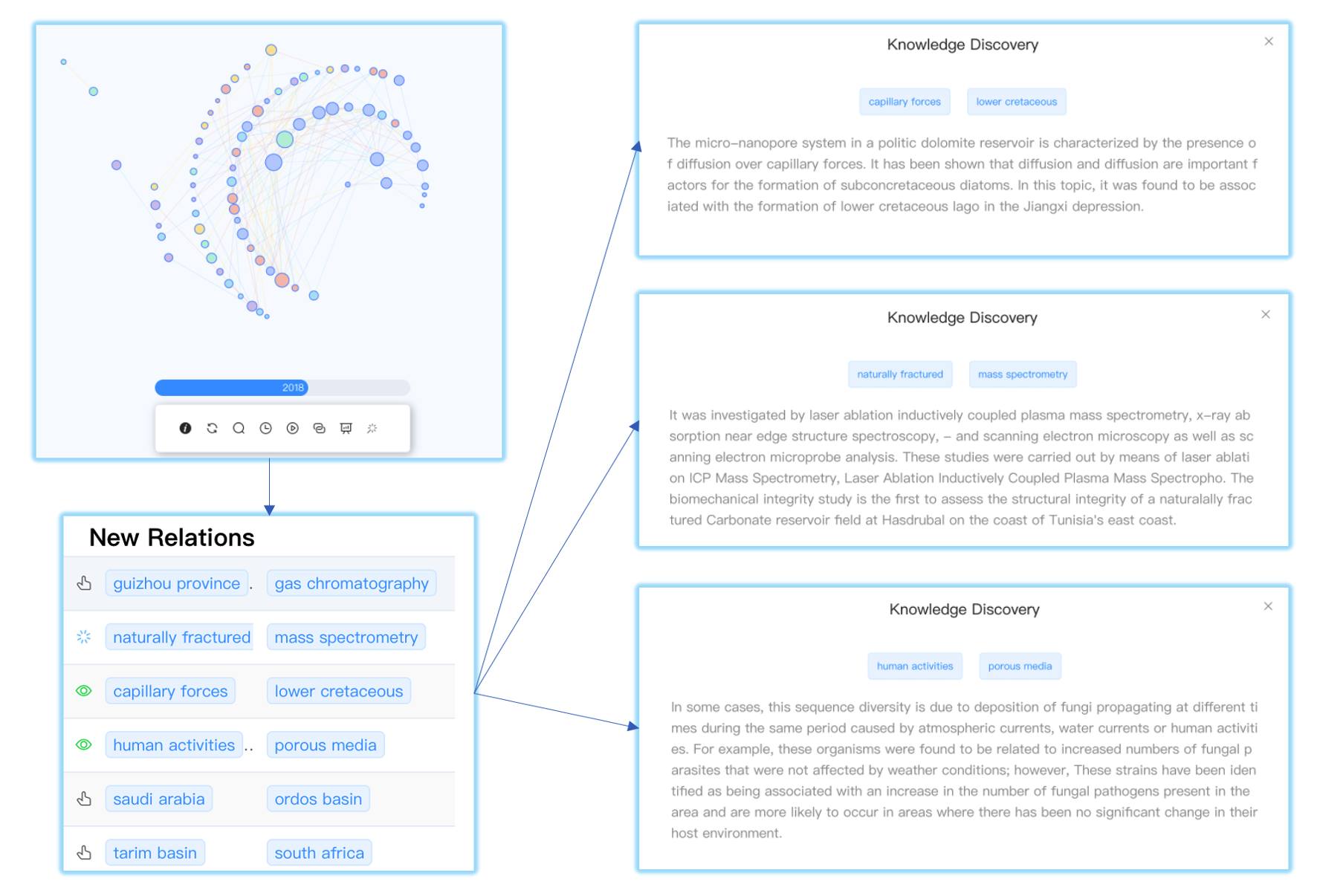}
    \caption{Screenshot of user interface: pipeline (public beta version).}
    \label{fig:screenshot2}
\end{figure*}

Our system (DeepReport) is available at website \url{https://idea.acemap.cn}. Figure \ref{fig:screenshot1} and \ref{fig:screenshot2} are screenshots of user interface (public beta version). As demonstrated in Figure \ref{fig:screenshot1}, after the concept "Carbonate Rock" is entered in the searching box, texts relevant to the keyword are presented in the insights box. The system will then dynamically construct an evolving concept co-occurrence graph based on the query result, where each node represents a concept, and relations between concepts are represented by the co-occurrence edges. We provide animations to demonstrate the evolution of the concept graph. The result of temporal link prediction is shown as concept pairs in the lower left \textit{New Relations} box, and verbalized idea for each pair is shown in a new dialog box. Researchers can select different concept pairs they are interested in and view the corresponding ideas, as illustrated in figure \ref{fig:screenshot2}. The system also provides network analytic tools such as community detection algorithms and Sankey diagrams for deeper investigation. The response time of the whole system is within 20 seconds.

\section{Potential Connections PLM-LP Predicted}\label{plmlppre}
We apply PLM-LP to the constructed 240 evolving concept co-occurrence graphs. We use all graph snapshots, including the year 2021, for training to mine potential connections that may appear in the future. We select the top $K$ pairs of concepts that are most likely to be connected by calculating the difference between the logits of labels, i.e., $``related"$ and $``unrelated"$. Table \ref{table:pre2022_result} presents potential connections PLM-LP predicted in 20 disciplines and topics. The connections are shown as concept pairs with \& concatenated. For each discipline, we only display six pairs as examples. In our human assessment, we recruited experts in the field of computer science and geo-science (geology and geography) to evaluate the predicted results in their corresponding domains. Their feedback reveals that at least a third of the potential concept pairs generated by the system are reasonable.

\section{Comparison Results of Link Predictions on All Disciplines}\label{cmp-lp-all}
PLM-LP is compared with three up-to-date temporal models: SEMNET~\cite{Krenn2020PredictingRT}, GCN-GAN~\cite{lei2019gcn}, and EvolveGCN~\cite{pareja2020evolvegcn}, which are applicable to the concept co-occurrence graph. In the experiment, their performance is evaluated on our constructed 240 concept co-occurrence graphs, where the last snapshot (the year 2021) is used as the test set. We report the accuracy of the adjacent matrix, precision, recall, and F1 score of all edges and new edges existing in the graph of 2021. New edges do not exist in the past years and would only come out in 2021. Results of these models in 20 disciplines/topics are provided in Table~\ref{table:subject_result}. It should be mentioned that we show the average of 12 evolving concept co-occurrence graphs of each discipline. The results show that GCN-GAN and EvolveGCN are unable to discover new edges. Our proposed PLM-LP is superior to any other models in the task of idea exploration, where the given graphs are strictly evolving network~\cite{skarding2021foundations}.

\section{Statistics of Evolving Concept Co-occurrence Graph}\label{statecc}
We construct 240 evolving concept co-occurrence graphs (12 graphs per discipline/topics) with Elasticsearch and Autophrase~\cite{shang2018automated} according to 240 essential and common queries and relevant papers. Each graph contains 22 temporal snapshots between 2000 and 2021. The statistics of the concept co-occurrence graphs are shown in Tables~\ref{table:query_stats1}, \ref{table:query_stats2}, \ref{table:query_stats3}, \ref{table:query_stats4}, and \ref{table:query_stats5}. These tables provide the corresponding discipline, query, number of nodes (concepts), number of edges in 2021, and selected concepts. We will release the construction code and data set on GitHub for further research, including temporal link prediction, community detection, academic trends analysis, knowledge representation, etc.

\section{About the Official Version of DeepReport}\label{officialversion}
In mid-2023, our DeepReport system underwent a major update, encompassing both data and model improvements. On the data front, we introduced a new version of the quintuple data (V202306), resulting in enhanced quality and a larger-scale dataset. The statistical summary of the new quintuple data (V202306) is presented in Table~\ref{table:quintuples202306}.

Furthermore, we trained a new state-of-the-art model in a specialized domain, which remains internal to our organization. This model, along with the integration of openAI's interface, was implemented to elevate the quality of our online services. The amalgamation of our proprietary large-scale model and the incorporation of openAI's resources empowered our system to provide superior performance and better cater to the needs of our users.

The introduction of the improved quintuple dataset, coupled with the deployment of the new specialized domain model and the utilization of openAI's interface, signifies a significant advancement in our DeepReport system. These updates enable us to deliver more accurate and reliable results, thereby enhancing the overall user experience. We remain committed to further refining our system to ensure it continues to meet the evolving demands of our users.

\section{Frequently Asked Questions}
\begin{itemize}
    \item \textbf{Q}: Comparing to other concepts graphs, what is the advantage of the concept co-occurrence quintuples? \textbf{A}: This question goes to the core of our work. This allows us to capture not only the co-occurrence relationship between concepts, but also their citation relationships, which can provide additional insights into how ideas are related (or generated) in academic literature. 

    \item \textbf{Q}: Why do you think transferring the concept links into natural languages is a necessary step in this assisting process? Your target users are academics. If they couldn't generate a proper idea from the link of concepts into natural language, do you expect the machines could do it better? \textbf{A}: The ultimate goal of our existing and future work is to enable LLM to generate reasonable, interpretable, and traceable ideas, and we now focus on how to use structured knowledge (here we use concept co-occurrence graphs) to guide this process. Therefore, the verbalization process is necessary. Besides, our system is designed to inspire researchers to discover ideas, rather than to replace them. Since we leverage the knowledge of 19 disciplines to train the model, it is valuable for researchers in certain cross-fields to broaden their points of interest.

    \item \textbf{Q}: The evaluation of verbalized ideas cannot measure logic, correctness, and details. Since you have already found the experts, could you think of some possibilities to evaluation the academic quality of these generated ideas trough human assessment? \textbf{A}: From our perspective, the human assessment of verbalized ideas is similar to the process of paper review but not identical, and the latter is more subjective. We should follow three principles, low resources, high efficiency, and objectivity to evaluate the generated ideas. For instance, predefined criteria and multiple rounds of voting can be of help. Meanwhile, we are trying to train human-in-the-loop-based system to assist experts in idea assessment.
\end{itemize}

\begin{table*}[]
\small
\begin{tabular}{|l|ll}
\hline
\textbf{Concept} & \multicolumn{2}{c|}{\textbf{plant disease \& machine learning}}\\ \hline
\textbf{Human}   & \multicolumn{2}{l|}{\begin{tabular}[c]{@{}l@{}}One potential application is the development of mobile disease diagnostics through machine learning \\ and crowdsourcing. Here we announce the release of over 50,000 expertly curated images on healthy and \\ infected leaves of crops plants through the existing online platform PlantVillage. We describe both the \\ data and the platform. These data are the beginning of an on-going, crowdsourcing effort to enable \\ computer vision approaches to help solve the problem of yield losses in crop plants due to infectious diseases.\end{tabular}} \\ \hline
\textbf{Machine} & \multicolumn{2}{l|}{\begin{tabular}[c]{@{}l@{}}We present a machine learning approach for disease detection in field images. It employs a deep CNN,\\  which is trained using a supervised learning approach. The proposed approach is able to detect the \\ presence of a number of different plant diseases in field images, including Fusarium wilt, Phytophthora \\ infestans, and Xanthomonas syringae. The results show that the model is robust to changes in lighting, \\ orientation, and crop species, and that it can be used in a real world application.\end{tabular}}\\ \hline
\textbf{Concept} & \multicolumn{2}{c|}{\textbf{network intrusion detection \& neural network}}  \\ \hline
\textbf{Human}   & \multicolumn{2}{l|}{\begin{tabular}[c]{@{}l@{}}An intelligent agent based distributed network intrusion detection system is presented. Compared with \\ the current network intrusion detection techniques, IA NIDS uses the parallel technique to reform and \\ detect the coming packet on the application layer, which widens the scale of intrusion detection. It uses \\ intelligent distributed mutual agent technique to enhance the ability of real time response and uses mix \\ detection method to improve the accuracy for detecting DDOS attacks.\end{tabular}} \\ \hline
\textbf{Machine} & \multicolumn{2}{l|}{\begin{tabular}[c]{@{}l@{}}We present a novel approach to network intrusion detection based on the use of a neural network model. \\ The neural network model is trained using a variety of anomaly detection techniques and anomaly \\ classification techniques. The model is able to detect a wide range of network intrusions, including denial \\ of service attacks, DDOS attacks, spoofing attacks, and a wide variety of worm attacks. The results \\ demonstrate the effectiveness of the approach by comparing it to a number of state of the art neural \\ network models, and by comparing the results with those of other neural network model based approaches.\end{tabular}}                                   \\ \hline
\textbf{Concept} & \multicolumn{2}{c|}{\textbf{metal catalyst \& metal nanoparticles}} \\ \hline
\textbf{Human}   & \multicolumn{2}{l|}{\begin{tabular}[c]{@{}l@{}}Carbon materials deliver a good opportunity to design catalysts for hydrodeoxygenation by providing \\ chemically inert surface and tunable functional groups to support the metal nanoparticles. The main \\ objective of the research is to give an account on the use of carbon material supported heterogeneous \\ catalysts for the HDO of lignin and related molecules for the synthesis of non petroleum chemical \\ products including hydrocarbons.The surface properties of the catalysts and mechanistic insights into \\ the conversion with key active sites are provided, which will help the designing of new and effective \\ catalysts for this biomass conversion.\end{tabular}} \\ \hline
\textbf{Machine} & \multicolumn{2}{l|}{\begin{tabular}[c]{@{}l@{}}The catalysts were characterized by X ray diffraction, scanning electron microscopy, Fourier transform \\ infrared spectroscopy, thermogravimetric analysis, and nitrogen adsorption–desorption isotherms. \\ The results showed that the catalysts exhibited high activity in the hydrodeoxygenation of lignin derived\\  bio oil under mild conditions. Moreover, the catalysts were also applied to the upgradation of bio oil derived\\  from the catalysis isoproanolysis in the organic phase. The high activity of the catalysts was attributed \\ to the synergistic effect of the metal nanoparticle.\end{tabular}}                                                                       \\ \hline
\end{tabular}
\caption{Examples of input concepts, human-written texts, and verbalized ideas according to our test dataset of quintuples.}
\label{table:quintuples_examples}
\end{table*}

\begin{table*}
        \small
	\begin{center}
            \setlength\tabcolsep{2pt}
		\begin{tabular}{|c|c|}
			\hline
			\textbf{Discipline} & \textbf{New Connections} \\
			\hline
			Art & \begin{tabular}[c]{p{20em}p{20em}}rogue taxidermy \& visual arts & claude cahun \& science fiction \\ 
  avant garde \& early paleozoic & zhuang zi \& wang guowei\\ 
  post modernism \& human environments  & west coast \& hip hop\end{tabular} \\ 
 \hline 
Biology & \begin{tabular}[c]{p{20em}p{20em}}spinal cord \& pancreatic cancer & grizzly bear \& gene flow\\ 
  arabidopsis thaliana \& heavy chain & splicing variants \& echinococcus granulosus \\ 
  rna interference\& body mass index RNA  & splicing variants \& echinococcus granulosus \end{tabular} \\ 
 \hline 
Business & \begin{tabular}[c]{p{20em}p{20em}}structural unemployment \& stock market & copyright law \& knowledge transfer\\ 
  industrial relations \& firm size & sale constraints \& macroeconomic variables \\ 
  economic growth \& greenhouse gas emissions  & subprime mortgage crisis \& IMF \end{tabular} \\ 
 \hline 
Chemistry  & \begin{tabular}[c]{p{20em}p{20em}}mass spectrometry \& aryl halides & phase transition \& density functional theory \\ 
  capillary electrophoresis \& optical rotation  & symmetry breaking \& hydrogen bond\\ 
  spinodal decomposition \& statistical mechanics  & canonical ensemble \& condensed matter \end{tabular} \\ 
 \hline 
Computer & \begin{tabular}[c]{p{20em}p{20em}}implicit bias \& biological inspiration & reading comprehension \& cognitive linguistics \\ 
  ambient intelligence \& information technology  & graph isomorphism \& ad hoc \\ 
  intrusion detection \& social network analysis  & game theory \& cognitive psychology\end{tabular} \\ 
 \hline 
Covid-19 & \begin{tabular}[c]{p{20em}p{20em}}alternative splicing \& medical genetics & proton pump inhibitors \& helicobacter pylori \\ 
  psoriatic arthritis \& life expectancy & allergic rhinitis \& hyperbaric oxygen \\ 
  serotonin syndrome \& herpes zoster  & immunologic memory \& rheumatic diseases \end{tabular} \\ 
 \hline 
Economics & \begin{tabular}[c]{p{20em}p{20em}}financial crisis \& pension plan  & credit default swap \& idiosyncratic volatility\\ 
  social justice \& wealth inequality & european union \& quantitative easing\\ 
  intellectual capital \& income distribution & quality management \& blockchain technology \end{tabular} \\ 
 \hline 
Engineering & \begin{tabular}[c]{p{20em}p{20em}}NLP \& collective intelligence  & kinetic energy \& stress relief\\ 
  finite element \& closed form & heat exchanger \& tip vortex \\ 
  neural network \& software reuse & wave propagation \& monte carlo\end{tabular} \\ 
 \hline 
Environmental Science  & \begin{tabular}[c]{p{20em}p{20em}}saginaw bay \& domestic sewage & lake victoria \& trophic state\\ 
  air pollutant \& night sky brightness & image segmentation \& stripe rust\\ 
  meridional overturning circulation \& solar activity  & electrostatic precipitator \& suspended matter \end{tabular} \\ 
 \hline 
Geography & \begin{tabular}[c]{p{20em}p{20em}}water resources \& conceptual framework  & ecosystem services \& ice sheet\\ 
  air pollution \& underground river & vadose zone \& loess plateau \\ 
  landsat thematic mapper \& dry seaso & pm2.5 concentrations \& ecological restoration \end{tabular} \\ 
 \hline 
Geology & \begin{tabular}[c]{p{20em}p{20em}}massive sulfide \& early carboniferous  & damping ratio \& hard rock\\ 
  rock mechanics \& laser scanning  & seismic hazard \& coal mining \\ 
  radioactive waste \& early cretaceous  & satellite imagery \& impact craters \end{tabular} \\ 
 \hline 
History & \begin{tabular}[c]{p{20em}p{20em}}public health \& economic growth & social movements \& cold war\\ 
  public service \& internet governance  & international law \& paradigm shift\\ 
  public finance \& environmental governance  & social security \& digital divide \end{tabular} \\ 
 \hline 
Materials Science  & \begin{tabular}[c]{p{20em}p{20em}}ion exchange \& aqueous solution & cathodic protection \& silicon dioxide\\ 
  barium titanate \& molecular sieve & electron microscope \& manganese dioxide \\ 
  pulsed laser deposition \& visible light  & thermal cycling \& finite difference \end{tabular} \\ 
 \hline 
Mathematics  & \begin{tabular}[c]{p{20em}p{20em}}computational fluid dynamics \& integral equation  & neural networks \& maximal matching \\ 
  heat transfer \& partial differential equations  & dynamical systems \& particle swarm optimization \\ 
  hubbard model \& phase velocity  & differential geometry \& heisenberg group \end{tabular} \\ 
 \hline 
Medicine & \begin{tabular}[c]{p{20em}p{20em}}breast cancer \& neural crest & clinical trials \& traditional chinese\\ 
  lactobacillus acidophilus \& bone mineral density  & femtosecond laser \& connective tissue\\ 
  drug repurposing \& genetic algorithm  & monoclonal antibody \& hair cell\end{tabular} \\ 
 \hline 
Philosophy & \begin{tabular}[c]{p{20em}p{20em}}logical positivism \& immanuel kant & filial piety \& critical thinking\\ 
  moral psychology \& traditional chinese  & economic philosophy \& higher education \\ 
  western philosophy \& ontological proof  & ontological proof \& volunteer activity \end{tabular} \\ 
 \hline 
Physics & \begin{tabular}[c]{p{20em}p{20em}}particle swarm optimizer \& pattern recognition  & quantum gravity \& baryon number\\ 
  neural networks \& quantum interference  & phase diagram \& wave vector\\ 
  neutron diffraction \& electric field & electric field \& ray tracing \end{tabular} \\ 
 \hline 
Political Science  & \begin{tabular}[c]{p{20em}p{20em}}conflict resolution \& cultural diplomacy  & media literacy \& public policy\\ 
  climate change \& civil society & foreign affairs \& granger causality \\ 
  gender equity \& economic crisis  & civic education \& participatory democracy \end{tabular} \\ 
 \hline 
Psychology  & \begin{tabular}[c]{p{20em}p{20em}}emotion regulation \& self awareness & prosocial behavior \& working memory\\ 
  family environment \& self concept & parahippocampal gyrus \& angelman syndrome \\ 
  chronic physical \& emotional disturbance  & williams syndrome \& frontal lobe \end{tabular} \\ 
 \hline 
Sociology  & \begin{tabular}[c]{p{20em}p{20em}}public policy \& sexual harassment & regional governance \& cultural heritage\\ 
  citizenship behaviors \& adult education  & middle class \& life satisfaction\\ 
  household income \& vocational education & opinion dynamics \& social exclusion \end{tabular} \\ 
 \hline

		\end{tabular}
	\end{center}
	\caption{Predicted connections of concepts in different disciplines.}
	\label{table:pre2022_result}
\end{table*}

\begin{table*}
	\begin{center}
		\setlength\tabcolsep{11pt}
            \scriptsize
            \renewcommand{\arraystretch}{0.5}
		\begin{tabular}{|l|c|c|ccc|ccc|}
			\hline
			\multirow{2}{*}{\textbf{Disciplines}} & \multirow{2}{*}{\textbf{Method}} & \multirow{2}{*}{\textbf{Accuracy}} & \multicolumn{3}{c|}{\textbf{All Edges in 2021}} &  \multicolumn{3}{c|}{\textbf{New Edges in 2021}}  \\
			\cline{4-9}
			 &  &  & \textbf{Precision} & \textbf{Recall} & \textbf{F1} & \textbf{Precision} & \textbf{Recall} & \textbf{F1} \\
			\hline
			\multirow{4}{*}{Art} & SEMNET & 0.454 & 0.075 & 0.484 & 0.116 & 0.003 & 0.533 & 0.006 \\
			& GCN-GAN & 0.985 & 1.000 & 0.891 & 0.941 & N/A & 0 & N/A \\
			& EvolveGCN & 0.998 & 1.000 & 0.984 & 0.992 & N/A & 0 & N/A \\
			& PLM-LP & 0.706 & 0.994 & 1.000 & 0.997 & 0.642 & 1.000 & 0.671 \\
			\hline
			\multirow{4}{*}{Biology} & SEMNET & 0.490 & 0.092 & 0.495 & 0.131 & 0.007 & 0.568 & 0.014 \\
			& GCN-GAN & 0.978 & 1.000 & 0.857 & 0.923 & N/A & 0 & N/A \\
			& EvolveGCN & 0.995 & 1.000 & 0.969 & 0.984 & N/A & 0 & N/A \\
			& PLM-LP & 0.834 & 0.972 & 0.999 & 0.983 & 0.675 & 0.953 & 0.691 \\
			\hline
			\multirow{4}{*}{Business} & SEMNET & 0.573 & 0.117 & 0.361 & 0.148 & 0.010 & 0.358 & 0.019 \\
			& GCN-GAN & 0.968 & 1.000 & 0.843 & 0.914 & N/A & 0 & N/A \\
			& EvolveGCN & 0.993 & 1.000 & 0.963 & 0.981 & N/A & 0 & N/A \\
			& PLM-LP & 0.766 & 0.979 & 1.000 & 0.989 & 0.521 & 1.000 & 0.538 \\
			\hline
			\multirow{4}{*}{Chemistry} & SEMNET & 0.424 & 0.106 & 0.654 & 0.175 & 0.008 & 0.660 & 0.015 \\
			& GCN-GAN & 0.968 & 1.000 & 0.840 & 0.913 & N/A & 0 & N/A \\
			& EvolveGCN & 0.994 & 1.000 & 0.970 & 0.985 & N/A & 0 & N/A \\
			& PLM-LP & 0.812 & 1.000 & 1.000 & 1.000 & 0.751 & 1.000 & 0.752 \\
			\hline
			\multirow{4}{*}{Computer Science} & SEMNET & 0.459 & 0.083 & 0.502 & 0.127 & 0.005 & 0.611 & 0.010 \\
			& GCN-GAN & 0.980 & 1.000 & 0.875 & 0.932 & N/A & 0 & N/A \\
			& EvolveGCN & 0.996 & 1.000 & 0.977 & 0.988 & N/A & 0 & N/A \\
			& PLM-LP & 0.593 & 0.993 & 1.000 & 0.996 & 0.383 & 1.000 & 0.426 \\
			\hline
			\multirow{4}{*}{Covid-19} & SEMNET & 0.378 & 0.059 & 0.617 & 0.098 & 0.005 & 0.689 & 0.010 \\
			& GCN-GAN & 0.979 & 1.000 & 0.796 & 0.882 & N/A & 0 & N/A \\
			& EvolveGCN & 0.995 & 1.000 & 0.947 & 0.973 & N/A & 0 & N/A \\
			& PLM-LP & 0.778 & 0.987 & 0.998 & 0.992 & 0.663 & 1.000 & 0.679 \\
			\hline
			\multirow{4}{*}{Economics} & SEMNET & 0.405 & 0.111 & 0.624 & 0.173 & 0.007 & 0.660 & 0.013 \\
			& GCN-GAN & 0.974 & 1.000 & 0.884 & 0.938 & N/A & 0 & N/A \\
			& EvolveGCN & 0.994 & 1.000 & 0.973 & 0.986 & N/A & 0 & N/A \\
			& PLM-LP & 0.629 & 0.852 & 0.997 & 0.910 & 0.246 & 0.941 & 0.275 \\
			\hline
			\multirow{4}{*}{Engineering} & SEMNET & 0.599 & 0.104 & 0.373 & 0.151 & 0.010 & 0.379 & 0.019 \\
			& GCN-GAN & 0.967 & 1.000 & 0.825 & 0.903 & N/A & 0 & N/A \\
			& EvolveGCN & 0.993 & 1.000 & 0.961 & 0.980 & N/A & 0 & N/A \\
			& PLM-LP & 0.757 & 0.959 & 1.000 & 0.977 & 0.513 & 1.000 & 0.545 \\
			\hline
			\multirow{4}{*}{Environmental Science} & SEMNET & 0.485 & 0.110 & 0.511 & 0.150 & 0.007 & 0.555 & 0.014 \\
			& GCN-GAN & 0.970 & 1.000 & 0.831 & 0.907 & N/A & 0 & N/A \\
			& EvolveGCN & 0.994 & 1.000 & 0.965 & 0.982 & N/A & 0 & N/A \\
			& PLM-LP & 0.714 & 0.956 & 1.000 & 0.975 & 0.451 & 0.998 & 0.470 \\
			\hline
			\multirow{4}{*}{Geography} & SEMNET & 0.521 & 0.086 & 0.495 & 0.129 & 0.005 & 0.514 & 0.009 \\
			& GCN-GAN & 0.981 & 1.000 & 0.884 & 0.938 & N/A & 0 & N/A \\
			& EvolveGCN & 0.996 & 1.000 & 0.979 & 0.989 & N/A & 0 & N/A \\
			& PLM-LP & 0.728 & 0.983 & 0.993 & 0.988 & 0.449 & 0.927 & 0.465 \\
			\hline
			\multirow{4}{*}{Geology} & SEMNET & 0.479 & 0.081 & 0.452 & 0.127 & 0.007 & 0.448 & 0.014 \\
			& GCN-GAN & 0.975 & 1.000 & 0.850 & 0.918 & N/A & 0 & N/A \\
			& EvolveGCN & 0.995 & 1.000 & 0.965 & 0.982 & N/A & 0 & N/A \\
			& PLM-LP & 0.758 & 0.998 & 1.000 & 0.999 & 0.622 & 1.000 & 0.641 \\
			\hline
			\multirow{4}{*}{History} & SEMNET & 0.566 & 0.111 & 0.464 & 0.150 & 0.005 & 0.496 & 0.009 \\
			& GCN-GAN & 0.983 & 1.000 & 0.894 & 0.944 & N/A & 0 & N/A \\
			& EvolveGCN & 0.997 & 1.000 & 0.980 & 0.990 & N/A & 0 & N/A \\
			& PLM-LP & 0.781 & 1.000 & 0.998 & 0.999 & 0.697 & 1.000 & 0.700 \\
			\hline
			\multirow{4}{*}{Materials Science} & SEMNET & 0.471 & 0.099 & 0.426 & 0.110 & 0.011 & 0.435 & 0.016 \\
			& GCN-GAN & 0.968 & 1.000 & 0.853 & 0.920 & N/A & 0 & N/A \\
			& EvolveGCN & 0.992 & 1.000 & 0.965 & 0.982 & N/A & 0 & N/A \\
			& PLM-LP & 0.618 & 0.900 & 1.000 & 0.940 & 0.252 & 1.000 & 0.291 \\
			\hline
			\multirow{4}{*}{Mathematics} & SEMNET & 0.489 & 0.106 & 0.477 & 0.166 & 0.006 & 0.448 & 0.011 \\
			& GCN-GAN & 0.974 & 1.000 & 0.888 & 0.940 & N/A & 0 & N/A \\
			& EvolveGCN & 0.995 & 1.000 & 0.979 & 0.990 & N/A & 0 & N/A \\
			& PLM-LP & 0.866 & 0.951 & 1.000 & 0.969 & 0.665 & 1.000 & 0.685 \\
			\hline
			\multirow{4}{*}{Medicine} & SEMNET & 0.474 & 0.108 & 0.541 & 0.168 & 0.007 & 0.537 & 0.014 \\
			& GCN-GAN & 0.970 & 1.000 & 0.849 & 0.917 & N/A & 0 & N/A \\
			& EvolveGCN & 0.994 & 1.000 & 0.971 & 0.985 & N/A & 0 & N/A \\
			& PLM-LP & 0.694 & 0.990 & 1.000 & 0.995 & 0.447 & 1.000 & 0.465 \\
			\hline
			\multirow{4}{*}{Philosophy} & SEMNET & 0.424 & 0.102 & 0.586 & 0.132 & 0.005 & 0.755 & 0.011 \\
			& GCN-GAN & 0.981 & 1.000 & 0.858 & 0.921 & N/A & 0 & N/A \\
			& EvolveGCN & 0.996 & 1.000 & 0.966 & 0.982 & N/A & 0 & N/A \\
			& PLM-LP & 0.586 & 0.985 & 0.984 & 0.985 & 0.423 & 1.000 & 0.439 \\
			\hline
			\multirow{4}{*}{Physics} & SEMNET & 0.512 & 0.120 & 0.629 & 0.186 & 0.012 & 0.618 & 0.023 \\
			& GCN-GAN & 0.973 & 1.000 & 0.893 & 0.943 & N/A & 0 & N/A \\
			& EvolveGCN & 0.993 & 1.000 & 0.974 & 0.987 & N/A & 0 & N/A \\
			& PLM-LP & 0.890 & 0.909 & 1.000 & 0.940 & 0.692 & 1.000 & 0.720 \\
			\hline
			\multirow{4}{*}{Political Science} & SEMNET & 0.424 & 0.106 & 0.552 & 0.167 & 0.005 & 0.545 & 0.010 \\
			& GCN-GAN & 0.976 & 1.000 & 0.865 & 0.926 & N/A & 0 & N/A \\
			& EvolveGCN & 0.996 & 1.000 & 0.975 & 0.987 & N/A & 0 & N/A \\
			& PLM-LP & 0.817 & 0.999 & 0.995 & 0.997 & 0.673 & 1.000 & 0.692 \\
			\hline
			\multirow{4}{*}{Psychology} & SEMNET & 0.495 & 0.112 & 0.565 & 0.162 & 0.008 & 0.623 & 0.016 \\
			& GCN-GAN & 0.978 & 1.000 & 0.864 & 0.926 & N/A & 0 & N/A \\
			& EvolveGCN & 0.994 & 1.000 & 0.966 & 0.983 & N/A & 0 & N/A \\
			& PLM-LP & 0.645 & 0.999 & 1.000 & 1.000 & 0.498 & 0.989 & 0.503 \\
			\hline
			\multirow{4}{*}{Sociology} & SEMNET & 0.445 & 0.099 & 0.567 & 0.160 & 0.005 & 0.613 & 0.011 \\
			& GCN-GAN & 0.976 & 1.000 & 0.867 & 0.928 & N/A & 0 & N/A \\
			& EvolveGCN & 0.996 & 1.000 & 0.975 & 0.987 & N/A & 0 & N/A \\
			& PLM-LP & 0.720 & 0.988 & 0.994 & 0.991 & 0.540 & 0.943 & 0.554 \\
			\hline
			\multirow{4}{*}{Average} & SEMNET & 0.478 & 0.099 & 0.519 & 0.146 & 0.007 & 0.552 & 0.013 \\
			& GCN-GAN & 0.975 & 1.000 & 0.860 & 0.924 & N/A & 0 & N/A \\
			& EvolveGCN & 0.995 & 1.000 & 0.970 & 0.985 & N/A & 0 & N/A \\
			& PLM-LP & 0.735 & 0.970 & 0.998 & 0.981 & 0.540 & 0.988 & 0.560 \\
			\hline
		\end{tabular}
	\end{center}
	\caption{Results of link prediction on different disciplines/topics. N/A means all cases have been predicted to be negative.}
	\label{table:subject_result}
\end{table*}

\begin{sidewaystable*}
	\begin{center}
            \tiny
            \renewcommand{\arraystretch}{1.15}
		\begin{tabular}{|l|l|l|l|l|}
			\hline
			\textbf{Discipline} & \textbf{Query} & \textbf{Num. of Nodes} & \textbf{Num. of Edges (2021)} & \textbf{Selected Concepts} \\
			\hline
			COVID-19 & How Is The Effectiveness Of Vaccines For COVID-19 & 106 & 815 & public health, clinical trial, infectious diseases \\
			\hline
			COVID-19 & How Many Variants Does COVID-19 Have? & 88 & 370 & amino acid, single nucleotide polymorphism, breast cancer \\
			\hline
			COVID-19 & What Do We Know About Asymptomatic Transmission Of COVID-19? & 110 & 642 & public health, polymerase chain reaction, united states \\
			\hline
			COVID-19 & What Are The Sequelae Of COVID-19? & 116 & 605 & logistic regression, odds ratio, united states \\
			\hline
			COVID-19 & Will The COVID-19 Vaccines And Boosters Work On The New Variants? & 125 & 724 & adverse event, clinical trials, haemophilus influenzae \\
			\hline
			COVID-19 & Antibodies And COVID-19 & 74 & 402 & monoclonal antibody, phage display, amino acid \\
			\hline
			COVID-19 & What Is The Difference Between COVID-19 And Influenza? & 110 & 1062 & public health, polymerase chain reaction, united states \\
			\hline
			COVID-19 & What To Do If You Come Into Close Contact With Someone With COVID-19 & 53 & 164 & public health, severe acute respiratory syndrome, united states \\
			\hline
			COVID-19 & Effective Ways To Prevent COVID-19 & 72 & 337 & public health, risk factor, health care \\
			\hline
			COVID-19 & Clinical Presentation Of Covid19 In Dementia Patients & 130 & 2034 & vascular dementia, frontotemporal dementia, mild cognitive impairment \\
			\hline
			COVID-19 & What Is The Effectiveness Of Drugs Being Developed To Treat COVID-19 Patients? & 128 & 727 & clinical trial, adverse event, united states \\
			\hline
			COVID-19 & \begin{tabular}[l]{@{}l@{}}What Is The Impact Of The Sars-Cov-2 (Covid19) Pandemic On The Morbidity And Mortality \\ Of High Risk Patients Undergoing Surgery\end{tabular} & 196 & 4382 & logistic regression, odds ratio, confidence interval \\
			\hline
			Computer Science & Local Community Detection With Hints & 68 & 236 & social network, complex network, computational complexity \\
			\hline
			Computer Science & Bias And Discrimination In Artificial Intelligence & 81 & 857 & artificial intelligence, artificial neural network, neural network \\
			\hline
			Computer Science & The Development Of Artificial Intelligence In China & 104 & 1156 & artificial intelligence, neural network, artificial neural network \\
			\hline
			Computer Science & Interpretability Of Artificial Intelligence & 94 & 1017 & artificial intelligence, neural network, artificial neural network \\
			\hline
			Computer Science & Current Trends Of Computer Graphics & 84 & 318 & user interface, virtual reality, graphical user interface \\
			\hline
			Computer Science & How To Improve The Application Of Machine Learning In Product Development & 124 & 1324 & machine learning, neural network, support vector machine \\
			\hline
			Computer Science & Commercialization Of Artificial Intelligence & 92 & 942 & artificial intelligence, neural network, artificial neural network \\
			\hline
			Computer Science & Natural Language Processing And Pretrained Model & 80 & 862 & natural language, natural language processing, machine translation \\
			\hline
			Computer Science & The Development Of Computer Vision & 89 & 494 & image processing, machine vision, visual acuity \\
			\hline
			Computer Science & The Development Of Graph Neural Network & 88 & 623 & neural network, artificial neural network, graph theory \\
			\hline
			Computer Science & Is Computer Science Considered Science Or Engineering? & 95 & 801 & mechanical engineering, life science, social sciences \\
			\hline
			Computer Science & How Will Artificial Intelligence Develop In The Future? & 106 & 1221 & artificial intelligence, neural network, artificial neural network \\
			\hline
			Geography & A Volcanic Eruption As The Earth’S Devastating Force & 134 & 1309 & volcanic eruption, volcanic ash, lava flow \\
			\hline
			Geography & Features And Qualities Of Coastal Erosion & 132 & 987 & remote sensing, climate change, sediment transport \\
			\hline
			Geography & Assessment Of Climate Sensitivity & 147 & 1826 & climate change, climate sensitivity, greenhouse gas \\
			\hline
			Geography & Geography And Economic Development & 92 & 940 & economic geography, economic development, economic growth \\
			\hline
			Geography & Impact Of Climate Change On Agrometeorological Disasters And Pests And Diseases & 152 & 2671 & natural disaster, global warming, food security \\
			\hline
			Geography & How Human Activities Contribute To Climate Change & 157 & 2688 & climate change, global warming, global climate change \\
			\hline
			Geography & Effect Of Ice Loss On Sea Level Rise & 151 & 3149 & sea level, sea level rise, climate change \\
			\hline
			Geography & Animal Extinction And Ways Of Preventing The Human Role In It & 107 & 718 & anxiety disorders, prefrontal cortex, medial prefrontal cortex \\
			\hline
			Geography & What Is The Impact Of Urban Expansion On Plant Diversity Change In Karst Regions Of Southwest China & 118 & 1288 & southwest china, climate change, south china \\
			\hline
			Geography & Assessment Method Of The Sea Turtle-Nesting Habitat Of Small Reef Islands & 144 & 1734 & sea turtle, green turtle, climate change \\
			\hline
			Geography & What Is The Impact Of Sea Level Rise On Ecological Infrastructure? & 167 & 1960 & climate change, storm surge, tide gauge \\
			\hline
			Geography & What Cause Short Term Sea Level Change In Cretaceous? & 176 & 2490 & climate change, late cretaceous, early cretaceous \\
			\hline
			Geology & Composition Of Geosphere & 91 & 383 & climate change, radioactive waste, global warming \\
			\hline
			Geology & Glacier Mass Change & 159 & 1859 & mass balance, climate change, digital elevation model \\
			\hline
			Geology & Evolution Of Sedimentary Rock Formation Of A Rock Association Level & 111 & 1749 & sedimentary rock, source rock, trace element \\
			\hline
			Geology & Superimposed Metamorphism Of Chinese Coal & 94 & 463 & coalbed methane, trace element, functional group \\
			\hline
			Geology & What Is The Complete Process Of Basin Formation? & 107 & 1414 & source rock, late cretaceous, early cretaceous \\
			\hline
			Geology & Effect Of The Combination Characteristics Of Rock Structural Plane On The Stability Of A Rock-Mass Slope & 53 & 328 & finite element, shear strength, open pit \\
			\hline
			Geology & How Do Geological Plates Change? & 109 & 980 & plate tectonics, north america, climate change \\
			\hline
			Geology & A Case Study Assessment Of Soil Liquefaction Potential & 68 & 323 & case histories, peak ground acceleration, shear wave velocity \\
			\hline
			Geology & Global Distribution Of Carbonate & 137 & 2127 & climate change, carbon cycle, carbon dioxide \\
			\hline
			Geology & What Is The Impact Of Man On Geo-Environment? & 111 & 582 & remote sensing, sustainable development, climate change \\
			\hline
			Geology & Differences In The Influence Of The Tectonic Setting Of The Earth On The Formation Of Magma & 109 & 1421 & rare earth element, rare earth, volcanic rock \\
			\hline
			Geology & Evolution Of Archean Continental Crust & 134 & 1511 & continental margin, oceanic crust, partial melting \\
			\hline
		\end{tabular}
	\end{center}
	\caption{Statistics of queries and corresponding evolving concept co-occurrence graphs in COVID-19, Computer Science, Geography, and Geology.}
	\label{table:query_stats1}
\end{sidewaystable*}

\begin{sidewaystable*}
	\begin{center}
            \tiny
            \renewcommand{\arraystretch}{1.15}
		\begin{tabular}{|l|l|l|l|l|}
			\hline
			\textbf{Discipline} & \textbf{Query} & \textbf{Num. of Nodes} & \textbf{Num. of Edges (2021)} & \textbf{Selected Concepts} \\
			\hline
			Mathematics & Birch And Swinnerton-Dyer Conjecture & 84 & 427 & betula pendula, growing season, silver birch \\
			\hline
			Mathematics & Topology And Differential Geometry & 82 & 1100 & differential geometry, differential equations, differential forms \\
			\hline
			Mathematics & What Is The Geometric Meaning For Rigidity In Riemannian Manifolds? & 54 & 382 & riemannian geometry, lie group, sectional curvature \\
			\hline
			Mathematics & Recent Developments In The Navier-Stokes Problem & 37 & 199 & finite element, fluid dynamics, reynolds number \\
			\hline
			Mathematics & Recent Developments In Riemann Hypothesis & 71 & 643 & riemann hypothesis, zeta function, riemann zeta function \\
			\hline
			Mathematics & When Is A Finite Type Dupin Hypersurface Of A Hypersphere Isoparametric? & 43 & 174 & euclidean space, riemannian manifold, vector field \\
			\hline
			Mathematics & Numerical Analysis And Scientific Computing & 127 & 2489 & scientific computing, numerical methods, numerical analysis \\
			\hline
			Mathematics & The Fundamental Group Of A Noncommutative Space & 60 & 643 & field theory, quantum mechanics, phase space \\
			\hline
			Mathematics & Double Phase Anisotropic Variational Problem & 75 & 471 & boundary condition, finite element, phase transition \\
			\hline
			Mathematics & Complex Network Analysis Of Nonlinear Time Series & 81 & 879 & neural network, artificial neural network, dynamical systems \\
			\hline
			Mathematics & P Versus Np Problem And It Approximation & 51 & 266 & approximation algorithm, combinatorial optimization, approximation ratio \\
			\hline
			Mathematics & Rationality Of Rigid Quiver Grassmannians & 38 & 227 & moduli space, lie algebra, fixed point \\
			\hline
			History & The Silk Road & 79 & 428 & silk road, central asia, bombyx mori \\
			\hline
			History & When Music Mattered? & 73 & 425 & music education, classical music, musical instrument \\
			\hline
			History & China And The West: Society And Culture & 94 & 349 & western culture, united states, cultural diversity \\
			\hline
			History & Governance In Ancient  & 59 & 225 & ancient greek, ancient greece, han dynasty \\
			\hline
			History & The Domestic Policy Of The European Weimar Republic & 125 & 1734 & european union, member state, czech republic \\
			\hline
			History & What Changed After The October Revolution? & 92 & 720 & october revolution, industrial revolution, united states \\
			\hline
			History & Capitalism In America & 98 & 1388 & latin america, latin american, united states \\
			\hline
			History & The Impact Of Maritime Trade On World Civilization. & 114 & 1289 & united states, free trade, east asia \\
			\hline
			History & The Reign And Life Of Queen Elizabeth & 93 & 600 & henry vi, elizabeth ii, edward vi \\
			\hline
			History & Governing The New Nation & 110 & 1022 & united states, united nations, case study \\
			\hline
			History & British Colonial Studies & 118 & 1145 & british empire, world war, united states \\
			\hline
			History & Social Movements In America & 142 & 2149 & social movement, united states, latin america \\
			\hline
			Psychology & Lesbian, Gay, Bisexual, Transgender & 54 & 489 & sexual orientation, mental health, united states \\
			\hline
			Psychology & Positive Psychology & 71 & 480 & positive psychology, mental health, organizational behavior \\
			\hline
			Psychology & Psychology And Criminology & 77 & 812 & criminal justice, social control, criminal behavior \\
			\hline
			Psychology & Personal Perception And Self-Consciousness & 62 & 247 & personality trait, self consciousness, college student \\
			\hline
			Psychology & The Suicide Intervention & 122 & 1835 & suicide prevention, mental health, public health \\
			\hline
			Psychology & Mental Health Of Children & 120 & 1779 & mental health, health care, mental illness \\
			\hline
			Psychology & How Does Cognitivism Differ From Behaviorism? & 50 & 95 & cognitive science, information processing, cultural differences \\
			\hline
			Psychology & Racism, Bias, And Discrimination & 81 & 684 & african american, united states, civil rights \\
			\hline
			Psychology & Does Group Polarization Affect The Minority? & 94 & 289 & united states, african american, health care \\
			\hline
			Psychology & Peer Pressure On Academic Performance & 93 & 799 & high school, peer group, peer pressure \\
			\hline
			Psychology & What Is The Role Of Cognitive Flexibility And Inhibition In Complex Dynamic Tasks & 101 & 744 & working memory, executive function, frontal cortex \\
			\hline
			Psychology & \begin{tabular}[l]{@{}l@{}}Sex Differences In Functional Connectivity Between Resting State Brain Networks \\ In Autism Spectrum Disorder \end{tabular} & 133 & 2477 & autism spectrum disorder, magnetic resonance imaging, functional magnetic resonance imaging \\
			\hline
			Economics & Globalization And Unemployment & 89 & 1022 & financial crisis, economic growth, united states \\
			\hline
			Economics & Supply Chain Management & 84 & 1023 & chain management, supply chain management, supply chain \\
			\hline
			Economics & Volatility And The Cross-Section Of Real Estate Equity Returns During Pandemic & 58 & 426 & real estate investment trust, financial crisis, asset allocation \\
			\hline
			Economics & Human Capital And China'S Future Growth & 91 & 1250 & economic growth, human capital, economic development \\
			\hline
			Economics & What Critical Approach To Neoclassical Economics Is Superior?  & 102 & 1255 & classical economics, economic theory, economic growth \\
			\hline
			Economics & The Economic Policy Uncertainty & 114 & 1218 & economic growth, monetary policy, united states \\
			\hline
			Economics & Us Earnings Inequality & 101 & 1024 & income inequality, united states, wage inequality \\
			\hline
			Economics & What Is The Application Of Fixed Point Theory In Financial And Economic Sciences? & 104 & 1015 & financial market, economic theory, financial crisis \\
			\hline
			Economics & The Digital Economy & 91 & 1001 & digital economy, digital divide, economic growth \\
			\hline
			Economics & How Rate Hikes Can Exacerbate Labor-Market Inequality & 96 & 1076 & income inequality, united states, human capital \\
			\hline
			Economics & Why Do Buyers Pay Different Prices For Comparable Products? & 74 & 202 & supply chain, online auction, information asymmetry \\
			\hline
			Economics & The Impact Of Globalization On Income Distribution In Emerging Economies & 124 & 2384 & income distribution, economic growth, developing countries \\
			\hline
		\end{tabular}
	\end{center}
	\caption{Statistics of queries and corresponding evolving concept co-occurrence graphs in Mathematics, History, Psychology, and Economics.}
	\label{table:query_stats2}
\end{sidewaystable*}

\begin{sidewaystable*}
	\begin{center}
            \tiny
            \renewcommand{\arraystretch}{1.15}
		\begin{tabular}{|l|l|l|l|l|}
			\hline
			\textbf{Discipline} & \textbf{Query} & \textbf{Num. of Nodes} & \textbf{Num. of Edges (2021)} & \textbf{Selected Concepts} \\
			\hline
			Sociology & Gender Inequality In America & 102 & 1534 & gender equality, latin america, economic growth \\
			\hline
			Sociology & Abortion And Abortion Rights & 95 & 1174 & united states, health care, reproductive rights \\
			\hline
			Sociology & Social Networks Addiction & 77 & 742 & social network, social networks, social support \\
			\hline
			Sociology & Does An Improvement In Rural Infrastructure Contribute To Alleviate Poverty In Pakistan? & 124 & 1741 & poverty reduction, developing countries, economic growth \\
			\hline
			Sociology & Spread Of False News & 66 & 188 & false alarm, social media, machine learning \\
			\hline
			Sociology & Regional Identity And Regional Change & 89 & 798 & european union, east asia, economic development \\
			\hline
			Sociology & What Is The Impact Of Intergenerational Mobility On Well-Being In Japan & 92 & 505 & united states, human capital, income inequality \\
			\hline
			Sociology & Social Media And Marketing & 100 & 1325 & social media, media market, social media marketing \\
			\hline
			Sociology & Can Populism Contribute To A More Inclusive Citizenship? & 97 & 785 & human rights, united states, global citizenship \\
			\hline
			Sociology & Advancement Of Women In Science & 113 & 963 & united states, national science foundation, career advancement \\
			\hline
			Sociology & How Are Functionalism And Conflict Theory Similar? & 80 & 333 & conflict resolution, conflict management, international relations \\
			\hline
			Sociology & How Do Social Networks Influence Educational Processes & 112 & 1119 & social network, social networks, social influence \\
			\hline
			Art & Modern Aesthetics & 47 & 155 & traditional chinese, chinese traditional, wang guowei \\
			\hline
			Art & Eastern Asian Art & 117 & 1174 & east asia, south asia, central asia \\
			\hline
			Art & Surrealist Aesthetics & 34 & 123 & andre breton, twentieth century, world war \\
			\hline
			Art & Modern Rock Music Trend & 88 & 956 & popular music, rock music, music industry \\
			\hline
			Art & Interaction Design & 83 & 736 & interaction design, interface design, user experience \\
			\hline
			Art & Victorian Beauty Standards In Art & 72 & 395 & victorian period, victorian era, victorian england \\
			\hline
			Art & Are Culturally Vibrant Communities Healthier? & 98 & 489 & united states, public health, cultural heritage \\
			\hline
			Art & The Role Of Cultural Identity In The Creation Of Art & 112 & 1002 & cultural identity, contemporary art, case study \\
			\hline
			Art & How Art Develops The Personality Of Human Beings? & 108 & 504 & human development, visual arts, personal development \\
			\hline
			Art & How Can Graffiti Be Accepted As A Form Of Street Art And Which Attributes Can Be Contributed To Architecture? & 79 & 492 & public space, street art, graffiti art \\
			\hline
			Art & The Use Of Arts Interventions For Mental Health And Wellbeing In Health Settings & 140 & 1695 & mental health, mental illness, public health \\
			\hline
			Art & What Is The Difference Between Islamic Art And Christian Art In Terms Of Function In The Middle Ages? & 126 & 1357 & middle age, middle east, islamic art \\
			\hline
			Business & Employee Motivation & 77 & 687 & job satisfaction, human resource, dependent variable \\
			\hline
			Business & International Trade Trends In The Usa & 113 & 1417 & developing countries, free trade, united states \\
			\hline
			Business & How Blockchain And Cryptocurrency Can Revolutionize Business? & 85 & 663 & business model, business process, blockchain technology \\
			\hline
			Business & What Is The Impact Of Intermediaries On A Negotiation? & 82 & 320 & united states, developing countries, european union \\
			\hline
			Business & International Business, Further Globalisation Or Backlash? & 117 & 949 & international trade, developing countries, business environment \\
			\hline
			Business & The Market Growth For Electric Vehicles & 92 & 1599 & electric vehicle, hybrid electric vehicle, hybrid electric \\
			\hline
			Business & Current Trends In Consumer Behavior & 74 & 330 & literature review, social media, online shopping \\
			\hline
			Business & \begin{tabular}[l]{@{}l@{}}Why Is The Importance Of The Correlation Analysis Between The Stock Market Valuation And The \\ Economic Situation Of Business Entities Growing? \end{tabular} & 100 & 1422 & stock market, financial market, capital market \\
			\hline
			Business & \begin{tabular}[l]{@{}l@{}}How To Evaluate Cost Impacts On Reverse Logistics Using An Economic Order Quantity (Eoq) Model With \\ Environmental And Social Considerations \end{tabular} & 61 & 330 & supply chain management, sustainable development, carbon emission \\
			\hline
			Business & Global Unemployment & 94 & 1085 & economic growth, united states, unemployment insurance \\
			\hline
			Business & Capitalism And Multinational Companies & 80 & 975 & multinational corporation, foreign direct investment, human capital \\
			\hline
			Business & Next Financial Crisis & 98 & 1897 & global financial crisis, financial market, financial crises \\
			\hline
			Physics & Microfluidics And Microsystems & 99 & 554 & cell culture, integrated circuit, sample preparation \\
			\hline
			Physics & How Swarm Robotics Can Be Used To Describe Particle System’S Deformation & 66 & 474 & particle swarm optimization, mobile robot, swarm intelligence \\
			\hline
			Physics & Dark Matter & 81 & 1208 & dark matter, direct detection, standard model \\
			\hline
			Physics & Galaxy Formation And Evolution & 91 & 1671 & star formation, galaxy evolution, galaxy formation \\
			\hline
			Physics & Big Bang (Quantum) Cosmology & 70 & 1092 & big bang, cosmological constant, dark matter \\
			\hline
			Physics & Global Nonlinear Stability Of Large Dispersive Solutions To The Einstein Equations & 55 & 318 & differential equation, cauchy problem, partial differential equations \\
			\hline
			Physics & How Does The Magnetoresistance Reflect The Information Of Fermi Surface? & 55 & 564 & fermi surface, magnetic field, fermi level \\
			\hline
			Physics & Condensed Matter Physics And Acoustics & 92 & 1156 & condensed matter, condensed matter physics, phase transition \\
			\hline
			Physics & The Developments In Quantum Computers & 92 & 1248 & quantum computing, quantum computers, quantum computation \\
			\hline
			Physics & Optical Physics And Quantum Information Science & 113 & 2216 & quantum information, quantum mechanics, quantum physics \\
			\hline
			Physics & The Space-Time Geometry Behind The Constant Speed Of Light & 118 & 1826 & general relativity, special relativity, cosmological constant \\
			\hline
			Physics & Antiferromagnetic Spintronic & 55 & 470 & magnetic field, magnetic moment, ground state \\
			\hline
		\end{tabular}
	\end{center}
	\caption{Statistics of queries and corresponding evolving concept co-occurrence graphs in Sociology, Art, Business, and Physics.}
	\label{table:query_stats3}
\end{sidewaystable*}

\begin{sidewaystable*}
	\begin{center}
            \tiny
            \renewcommand{\arraystretch}{1.15}
		\begin{tabular}{|l|l|l|l|l|}
			\hline
			\textbf{Discipline} & \textbf{Query} & \textbf{Num. of Nodes} & \textbf{Num. of Edges (2021)} & \textbf{Selected Concepts} \\
			\hline
			Political Science & Politics And Diplomacy & 94 & 1261 & foreign policy, public diplomacy, united states \\
			\hline
			Political Science & Democracy And The Public In The European Union & 132 & 2406 & european union, member state, european integration \\
			\hline
			Political Science & What Is The Most Powerful Act Of Political Participation? & 101 & 795 & united states, political parties, civil society \\
			\hline
			Political Science & Rural Revitalization & 108 & 583 & economic development, sustainable development, united states \\
			\hline
			Political Science & Opportunities And Challenges Facing China’S Economic “External Circulation” & 94 & 788 & economic development, economic growth, sustainable development \\
			\hline
			Political Science & How Politicians Use Social Media? & 79 & 891 & social media, social network, mass media \\
			\hline
			Political Science & The Basics Of The Theoretical System Of Socialism With Chinese Characteristics & 83 & 576 & chinese character, communist party, deng xiaoping \\
			\hline
			Political Science & \begin{tabular}[l]{@{}l@{}}How Political Orientation, Economic Precarity, And Participant Demographics Impact Compliance With \\ COVID-19 Prevention Measures In A Dutch Representative Sample? \end{tabular} & 119 & 737 & economic development, united states, economic growth \\
			\hline
			Political Science & Chinese Communist Party Hierarchy & 67 & 509 & chinese communist party, communist party, chinese communist \\
			\hline
			Political Science & Land System Reformation & 74 & 681 & land reform, land tenure, agrarian reform \\
			\hline
			Political Science & Vietnam War Interests Aggregation & 116 & 2016 & vietnam war, united states, south vietnam \\
			\hline
			Political Science & Russia And Nato Relationships & 113 & 2671 & cold war, united states, north atlantic \\
			\hline
			Philosophy & Self Consciousness & 52 & 183 & altered states, global workspace theory, vegetative state \\
			\hline
			Philosophy & Why Do We Strive For Perfection If It Is Not Attainable? & 26 & 74 & higher education, united states, human capital \\
			\hline
			Philosophy & Dealistic Understanding Of Existence & 41 & 68 & conceptual model, conceptual framework, neural network \\
			\hline
			Philosophy & Philosophical Anthropology & 49 & 252 & philosophical anthropology, cultural anthropology, medical anthropology \\
			\hline
			Philosophy & Research On Gadamer'S Philosophy & 36 & 178 & social science, western philosophy, chinese philosophy \\
			\hline
			Philosophy & Metaprobes, Metaphysical And Sketchy Philosophy & 59 & 319 & western philosophy, modern philosophy, greek philosophy \\
			\hline
			Philosophy & Why Is Beauty Associated With Morality? & 55 & 179 & moral philosophy, human nature, university press \\
			\hline
			Philosophy & Apocalypse And The Ends Of The World & 50 & 205 & united states, world war ii, cold war \\
			\hline
			Philosophy & Cultural Genesis And Dynamics Of Culture & 68 & 295 & cultural diversity, cultural identity, national culture \\
			\hline
			Philosophy & Is Confucianism A Religious Philosophy Or Ethics & 86 & 612 & chinese philosophy, moral philosophy, human nature \\
			\hline
			Philosophy & Where Does Your Self-Worth Come From? & 65 & 289 & mental health, high school, body image \\
			\hline
			Philosophy & Where Do You Find Meaning In Your Life? & 54 & 238 & mental health, college students, regression analysis \\
			\hline
			Biology & Endangered Species Recovery & 77 & 615 & endangered species, united states, critically endangered \\
			\hline
			Biology & Gene Modification And Disease & 131 & 1791 & gene expression, histone modification, gene therapy \\
			\hline
			Biology & Crispr And Genetic Engineering & 79 & 569 & genetic engineering, genetically engineered, genetically modified \\
			\hline
			Biology & The Effect Of Plant Genome Editing & 91 & 1059 & arabidopsis thaliana, gene expression, rna editing \\
			\hline
			Biology & Biosynthesis, Transport And Biological Functions Of Ascorbic Acid In Plants. & 130 & 1537 & ascorbic acid, amino acid, arabidopsis thaliana \\
			\hline
			Biology & How To Estimate Female Malaria Mosquito Age By Quantifying Y-Linked Genes In Stored Male Spermatozoa & 114 & 773 & public health, aedes aegypti, yellow fever \\
			\hline
			Biology & Evolution Of Terrestrial Plant  & 94 & 502 & climate change, fossil record, carbon dioxide \\
			\hline
			Biology & The Effect Of Synergistic Interaction Between Earthworms And Microorganisms On The Composting Process. & 95 & 638 & solid waste, sewage sludge, heavy metal \\
			\hline
			Biology & \begin{tabular}[l]{@{}l@{}}Development Of A Filter Device For The Prevention Of Aquatic Bacterial Disease Using A Single-Chain \\ Variable Fragment (Scfv)-Conjugated Affinity Silk \end{tabular} & 121 & 780 & western blot, polymerase chain reaction, flow cytometry \\
			\hline
			Biology & Human Cdna Clones & 103 & 848 & open reading frame, polymerase chain reaction, cell line \\
			\hline
			Biology & Interactions Between Genes & 131 & 959 & single nucleotide polymorphism, logistic regression, dimensionality reduction \\
			\hline
			Biology & Aging, Lifespan And Metabolism & 131 & 2219 & oxidative stress, gene expression, dietary restriction \\
			\hline
			Medicine & Development Of Xenotransplantation & 64 & 465 & stem cell, clinical trials, cell line \\
			\hline
			Medicine & Neuronal Regeneration & 127 & 1962 & spinal cord, dorsal root, stem cell \\
			\hline
			Medicine & The Role Of Vitamin D In The Pathogenesis Of Allergic Rhinitis & 117 & 1084 & mast cell, immune response, atopic dermatitis \\
			\hline
			Medicine & Tumor Immunity And Targeted Therapy & 119 & 2965 & immune response, tumor microenvironment, dendritic cell \\
			\hline
			Medicine & Early Detection Of Cancer & 109 & 1019 & breast cancer, colorectal cancer, prostate cancer \\
			\hline
			Medicine & Is Medical Research On Animals Ethical & 108 & 1642 & medical ethics, research ethics, ethics committee \\
			\hline
			Medicine & Association Between Migraine And Risk Of Ocular Motor Cranial Nerve Palsy & 127 & 2227 & cranial nerve, case report, magnetic resonance imaging \\
			\hline
			Medicine & Intestinal Flora Correlates With Chronic Liver Disease & 176 & 2798 & alcoholic liver disease, nonalcoholic fatty liver disease, hepatocellular carcinoma \\
			\hline
			Medicine & Treatment Of Alzheimer'S Disease & 128 & 1146 & clinical trials, cognitive function, animal model \\
			\hline
			Medicine & Medical Humanitarian Missions In The Developing World & 113 & 1790 & humanitarian assistance, humanitarian aid, united nations \\
			\hline
			Medicine & Artificial Intelligence In Vaccine And Drug Design & 99 & 1058 & artificial intelligence, neural network, machine learning \\
			\hline
			Medicine & What The Effect Of Weightbearing And Foot Positioning On 3D Distal Tibiofibular Joint Parameters? & 67 & 385 & soft tissue, computed tomography, sagittal plane \\
			\hline
		\end{tabular}
	\end{center}
	\caption{Statistics of queries and corresponding evolving concept co-occurrence graphs in Political Science, Philosophy, Biology, and Medicine.}
	\label{table:query_stats4}
\end{sidewaystable*}

\begin{sidewaystable*}
	\begin{center}
            \tiny
            \renewcommand{\arraystretch}{1.15}
		\begin{tabular}{|l|l|l|l|l|}
			\hline
			\textbf{Discipline} & \textbf{Query} & \textbf{Num. of Nodes} & \textbf{Num. of Edges (2021)} & \textbf{Selected Concepts} \\
			\hline
			Materials Science & Interconversion Of Multiferroic Domains And Domain Walls & 55 & 320 & electric field, magnetic fields, room temperature \\
			\hline
			Materials Science & Topological Insulators & 49 & 260 & magnetic field, hall effect, band structure \\
			\hline
			Materials Science & Programmable Matter & 93 & 234 & south africa, higher education, health care \\
			\hline
			Materials Science & Selective Laser Sintering & 75 & 486 & additive manufacturing, rapid prototyping, scanning electron microscopy \\
			\hline
			Materials Science & Perovskite Ferroelectric Material & 60 & 813 & phase transition, room temperature, electric field \\
			\hline
			Materials Science & High-Temperature Superconductivity & 93 & 981 & critical temperature, magnetic field, current density \\
			\hline
			Materials Science & Electromagnetic Wave Absorbing Material & 70 & 600 & electromagnetic wave, electromagnetic interference, electromagnetic field \\
			\hline
			Materials Science & Unleaded Energy Storage Ceramics & 63 & 463 & energy storage, renewable energy, energy density \\
			\hline
			Materials Science & Preparation Of Composite Structures Of Titanium Dioxide Nanotube Arrays. & 123 & 1876 & electron microscopy, scanning electron microscopy, titanium oxide \\
			\hline
			Materials Science & \begin{tabular}[l]{@{}l@{}}Effect Of The Grain Arrangements On The Thermal Stability Of Polycrystalline Nickel-Rich \\ Lithium-Based Battery Cathodes \end{tabular} & 94 & 1566 & lithium ion, thermal stability, energy density \\
			\hline
			Materials Science & Erythritol Phase Change Thermal Storage Subcooling And Thermal Conductivity Improvement & 95 & 1839 & thermal conductivity, phase change, thermal energy storage \\
			\hline
			Materials Science & The Development Of Nanomaterials & 89 & 595 & carbon nanotube, quantum dot, metal oxide \\
			\hline
			Environmental Science & Noise And Light Pollution & 85 & 664 & noise pollution, air pollution, environmental noise \\
			\hline
			Environmental Science & Microplastic Impacts On Ecosystem & 120 & 1029 & food web, climate change, trophic level \\
			\hline
			Environmental Science & Removal Of Refractory Organic Pollutants & 84 & 866 & organic compounds, wastewater treatment, organic matter \\
			\hline
			Environmental Science & Remote Sensing And Geographic Information Systems & 112 & 845 & remote sensing, image processing, change detection \\
			\hline
			Environmental Science & Influence Of Hydrodynamics On Nutrient Cycling And Algal Growth In Taihu Lake & 97 & 907 & water quality, meiliang bay, organic matter \\
			\hline
			Environmental Science & Atmospheric Environmental Capacity Accounting And Total Pollutant Control & 106 & 1252 & air pollution, air quality, air pollutants \\
			\hline
			Environmental Science & Environmental Science And Sustainable Development & 135 & 2648 & sustainable development, environmental sustainability, environmental science \\
			\hline
			Environmental Science & Generation And Direct Observation Of A Triplet Arylnitrenium Ion & 69 & 510 & excited state, ground state, electron transfer \\
			\hline
			Environmental Science & What Are The Goals Of Environmental Science Studies? & 126 & 1894 & environmental science, case study, sustainable development \\
			\hline
			Environmental Science & Why Chemists Can’T Quit Palladium & 51 & 216 & room temperature, organic synthesis, transition metal \\
			\hline
			Environmental Science & Global Warming And Climate Change & 141 & 2287 & global warming, greenhouse gas, carbon dioxide \\
			\hline
			Environmental Science & Mercury Pollution Elimination & 109 & 823 & heavy metal, air pollution, food chain \\
			\hline
			Chemistry & Photoelectrochemical Biosensor & 102 & 853 & surface plasmon resonance, quantum dot, glucose oxidase \\
			\hline
			Chemistry & Axial Chiral Compounds & 48 & 314 & optically active, amino acid, circular dichroism \\
			\hline
			Chemistry & Organic Chemistry And Discovery & 73 & 892 & organic chemist, drug discovery, organic synthesis \\
			\hline
			Chemistry & Nitrogen Heterocyclic Carbene Catalysis & 44 & 286 & transition metal, homogeneous catalysis, room temperature \\
			\hline
			Chemistry & Noncovalent Interaction & 89 & 1087 & hydrogen bond, hydrogen bonding, density functional \\
			\hline
			Chemistry & Hydrophobic Effect Phenomenon & 117 & 947 & contact angle, aqueous solution, hydrogen bond \\
			\hline
			Chemistry & Colloid Theory & 84 & 688 & ionic strength, light scattering, porous media \\
			\hline
			Chemistry & How To Make A Fruitier, More Floral Chocolate & 74 & 308 & cocoa butter, fatty acid, cocoa bean \\
			\hline
			Chemistry & Organic Chemical Reactivity Functioning & 85 & 743 & functional group, organic compound, density functional theory \\
			\hline
			Chemistry & Why Do Transition Crystals (Hybrid Crystals) Conduct Electricity? & 82 & 1084 & electrical conductivity, single crystal, phase transition \\
			\hline
			Chemistry & Can Electric Fields Drive Chemistry For An Aqueous Microdroplet? & 88 & 432 & electric field, aqueous solution, magnetic field \\
			\hline
			Chemistry & What Are The Downstream Products Generated From Coal? & 105 & 1088 & carbon dioxide, natural gas, bituminous coal \\
			\hline
			Engineering & Flexible Surgical Robot & 64 & 267 & surgical instrument, da vinci, laparoscopic surgery \\
			\hline
			Engineering & Insect Like Micro Air Vehicle & 71 & 328 & wind tunnel, air pollution, electric vehicles \\
			\hline
			Engineering & The Use Of Ai And Machine Learning In Engineering & 124 & 2009 & artificial intelligence, machine learning, neural network \\
			\hline
			Engineering & Civil Engineering  & 85 & 448 & civil engineer, structural engineering, environmental engineering \\
			\hline
			Engineering & Aerodynamics And Fluid Mechanics & 74 & 949 & computational fluid dynamics, fluid dynamics, wind tunnel \\
			\hline
			Engineering & Crowdsourcing In Software Engineering & 101 & 1098 & software engineer, software development, requirements engineering \\
			\hline
			Engineering & The Effect Of Stress Release On The Stability Of Excavation Works & 34 & 130 & power station, finite element, stress concentration \\
			\hline
			Engineering & Specifics Of Engineering Materials & 103 & 557 & composite material, materials science, civil engineering \\
			\hline
			Engineering & Risk Caused By The Propagation Of Earthquake Losses Through The Economy & 87 & 633 & risk assessment, risk management, seismic hazard \\
			\hline
			Engineering & Heat Transfer In Low Temperature & 89 & 1446 & heat transfer, heat transfer coefficient, heat flux \\
			\hline
			Engineering & Effect Of Piston Structural Stiffness On Dynamic Performance & 60 & 259 & finite element, finite element analysis, finite element method \\
			\hline
			Engineering & Effects Of Thickness Reduction In Cold Rolling Process On The Formability Of Sheet Metals Using Anfis & 70 & 673 & room temperature, grain size, heat treatment \\
			\hline
		\end{tabular}
	\end{center}
	\caption{Statistics of queries and corresponding evolving concept co-occurrence graphs in Materials Science, Environmental Science, Chemistry, and Engineering.}
	\label{table:query_stats5}
\end{sidewaystable*}

\begin{table*}
	\begin{center}
        \renewcommand{\arraystretch}{1.5}
		\begin{tabular}{|l|c|c|c|c|c|}
			\hline
			\textbf{Discipline} & \textbf{Quintuple} & \textbf{Concept} & \textbf{Concept Pair} & \textbf{Total $p$} & \textbf{Total $p_1$ \& $p_2$} \\
			\hline
			Art & 7,510 & 2,671 & 5,845 & 2,770 & 7,060 \\
			\hline
			History & 5,287 & 2,198 & 4,654 & 2,348 & 5,764 \\
			\hline
			Philosophy & 45,752 & 4,773 & 25,935 & 16,896 & 29,942 \\
			\hline
			Sociology & 16,017 & 4,054 & 12,796 & 7,066 & 16,416 \\
			\hline
			Political Science & 67,975 & 6,105 & 42,411 & 26,198 & 53,933 \\
			\hline
			Business & 205,297 & 9,608 & 99,329 & 62,332 & 112,736 \\
			\hline
			Geography & 191,958 & 12,029 & 118,563 & 42,317 & 112,909 \\
			\hline
			Engineering & 506,635 & 16,992 & 249,935 & 137,164 & 273,894 \\
			\hline
			Geology & 365,183 & 13,795 & 190,002 & 98,991 & 222,358 \\
			\hline
			Medicine & 168,697 & 13,014 & 114,104 & 42,535 & 138,973 \\
			\hline
			Economics & 227,530 & 9,461 & 113,527 & 68,607 & 131,387 \\
			\hline
			Physics & 267,532 & 10,831 & 133,079 & 84,824 & 176,741 \\
			\hline
			Biology & 224,722 & 15,119 & 145,088 & 59,210 & 189,281 \\
			\hline
			Mathematics & 312,670 & 17,751 & 190,734 & 95,951 & 218,697 \\
			\hline
			Psychology & 476,342 & 9,512 & 194,038 & 115,725 & 212,180 \\
			\hline
			Computer Science & 531,654 & 16,591 & 244,567 & 151,809 & 238,091 \\
			\hline
			Environmental Science & 583,466 & 11,002 & 226,671 & 94,474 & 201,330 \\
			\hline
			Materials Science & 573,032 & 17,098 & 249,251 & 145,068 & 313,657 \\
			\hline
			Chemistry & 565,307 & 13,858 & 231,062 & 108,637 & 286,593 \\
			\hline
			\textbf{Total} & 5,342,566 & 206,462 & 2,591,591 & 1,362,922 & 2,941,942 \\
			\hline
		\end{tabular}
	\end{center}
	\caption{Statistics of Quintuples V202306}
	\label{table:quintuples202306}
\end{table*}

\end{document}